\documentclass{article}
\usepackage{graphicx}
\usepackage{subcaption}
\usepackage[margin=1in]{geometry}
\usepackage{amsmath}
\usepackage{multirow}
\usepackage{amssymb}
\usepackage{cite}
\usepackage{booktabs}
\usepackage{float}
\usepackage{tikz}
\usepackage{authblk}
\usepackage{algorithm}
\usepackage{algorithmic}
\usepackage{url}
\usetikzlibrary{positioning, calc, arrows.meta}

\title{Adapting Video Diffusion Models for Time-Lapse Microscopy}

\author{Alexander Holmberg}
\author{Nils Mechtel}
\author{Wei Ouyang\thanks{Correspondence: \texttt{weio@kth.se}}}

\affil{Department of Applied Physics, Science for Life Laboratory, 
          KTH Royal Institute of Technology, Stockholm, Sweden}
\date{\today}

\begin{document}

\maketitle

\begin{abstract} 
We present a domain adaptation of video diffusion models to generate highly realistic time-lapse microscopy videos of cell division in HeLa cells. Although state-of-the-art generative video models have advanced significantly for natural videos, they remain underexplored in microscopy domains. To address this gap, we fine-tune a pretrained video diffusion model on microscopy-specific sequences, exploring three conditioning strategies: (1) text prompts derived from numeric phenotypic measurements (e.g., proliferation rates, migration speeds, cell-death frequencies), (2) direct numeric embeddings of phenotype scores, and (3) image-conditioned generation, where an initial microscopy frame is extended into a complete video sequence. Evaluation using biologically meaningful morphological, proliferation, and migration metrics demonstrates that fine-tuning substantially improves realism and accurately captures critical cellular behaviors such as mitosis and migration. Notably, the fine-tuned model also generalizes beyond the training horizon, generating coherent cell dynamics even in extended sequences. However, precisely controlling specific phenotypic characteristics remains challenging, highlighting opportunities for future work to enhance conditioning methods. Our results demonstrate the potential for domain-specific fine-tuning of generative video models to produce biologically plausible synthetic microscopy data, supporting applications such as in-silico hypothesis testing and data augmentation.
\end{abstract}

\section{Introduction} \label{sec:intro}
Time-lapse microscopy is essential for studying dynamic cellular processes such as division, migration, and morphological changes. Observations of these processes in widely-used cell lines, such as HeLa cells, can reveal how genetic perturbations influence critical cellular behaviors, including mitosis and proliferation. However, state-of-the-art generative video models trained primarily on natural video datasets struggle to capture the subtle morphological and temporal characteristics unique to microscopy sequences. To address this limitation, we adapt Video Diffusion Models—specifically CogVideoX~\cite{yang2024cogvideoxtexttovideodiffusionmodels}—for generating realistic microscopy videos depicting mitotic events. Given that zero-shot approaches typically fail to reproduce biologically accurate dynamics, we fine-tune the pretrained model on domain-specific microscopy data. This significantly enhances the realism and coherence of the generated videos. To quantitatively assess the realism in generated videos, we use biologically meaningful metrics that can be put into two categories:
\begin{enumerate}
\item \textbf{Morphological metrics}, which quantify key nuclear shape properties including area, eccentricity, perimeter, and solidity.
\item \textbf{Proliferation and migration metrics}, capturing dynamic behaviors such as cell-division events, growth ratios, final cell counts, and cell displacement.
\end{enumerate}

We examine different conditioning methods to guide video generation effectively. We implement a text-based approach, converting numeric phenotype measurements (e.g., proliferation rates, migration speeds, cell-death frequencies) into visually intuitive textual prompts, as well as numeric embedding strategies directly encoding phenotype scores into the model’s latent space.  We additionally investigate image-conditioned generation, where the model synthesizes sequences from real initial microscopy frames. Beyond simply creating visually compelling videos, such generative models have practical utility in in silico hypothesis testing and data augmentation. For instance, realistic synthetic time-lapse data could be used to pre-train or stress-test cell-tracking pipelines under a wider variety of conditions (e.g., lower cell density, abnormal mitotic rates) than might be available in real-world datasets. In this way, researchers could systematically explore how tracking or segmentation algorithms perform on rare or extreme phenotypes, potentially revealing their strengths and limitations before expensive lab experiments

Our experiments demonstrate substantial improvements in video realism over zero-shot baselines, indicating that the adapted models successfully internalize critical biological patterns. We also find that the fine-tuned model generalizes beyond the 81-frame training horizon, generating coherent cellular dynamics in extended sequences. While explicit phenotype conditioning via textual or numeric embeddings remains challenging, our results highlight the significant potential of domain-specific fine-tuning for generating biologically plausible microscopy time-lapse videos. Our contributions are threefold: (1) we introduce domain adaptation of Video Diffusion Models for time-lapse microscopy, (2) we propose phenotype-based conditioning via textual and numeric embeddings, and (3) we evaluate the generated videos with biologically specific metrics.

\section{Related Work}

\paragraph{Video Diffusion Models}  
Diffusion models transform Gaussian noise gradually into coherent data samples by learning the reverse of a predefined forward noising process. Formally, given a data sample (e.g., an image) \(\mathbf{x}_0\), a forward process progressively adds noise over \(T\) steps:
\begin{equation}
q(\mathbf{x}_t|\mathbf{x}_{t-1}) = \mathcal{N}(\mathbf{x}_t; \sqrt{1 - \beta_t}\,\mathbf{x}_{t-1},\,\beta_t \mathbf{I}), \quad t = 1,\dots,T,
\end{equation}
resulting in \(\mathbf{x}_T \sim \mathcal{N}(\mathbf{0},\mathbf{I})\). The reverse process is then learned by a parameterized neural network \(\boldsymbol{\theta}\):

\begin{equation}
p_{\boldsymbol{\theta}}(\mathbf{x}_{t-1}|\mathbf{x}_t) = \mathcal{N}(\mathbf{x}_{t-1}; \boldsymbol{\mu}_{\boldsymbol{\theta}}(\mathbf{x}_t, t), \boldsymbol{\Sigma}_{\boldsymbol{\theta}}(\mathbf{x}_t, t)).
\end{equation}
When extending diffusion models to video data, each sample becomes a 4D tensor \(\mathbf{x}_0 \in \mathbb{R}^{C \times F \times H \times W}\), where \(C\), \(F\), \(H\), and \(W\) correspond respectively to channels, frames, height, and width. Early approaches operated in pixel space and factorized time and space dimensions \cite{ho2022videodiffusionmodels, ho2022imagenvideohighdefinition}, employing a (2+1)D attention mechanism. Blattman et al. subsequently introduced latent video diffusion \cite{blattmann2023alignlatentshighresolutionvideo}, enabling a less computationally intensive training process. Recent models like CogVideoX \cite{yang2024cogvideoxtexttovideodiffusionmodels} and HunyuanVideo \cite{kong2025hunyuanvideosystematicframeworklarge} take advantage of the computational improvement of operating in a latent space and adopt a full 3D spatiotemporal attention. While these models achieve strong results on generating natural videos, they generally struggle with domain-specific content, such as microscopy videos, because they are primarily trained on natural video datasets.

\paragraph{Image Diffusion Models for Microscopy Datasets}
MorphoDiff \cite{Navidi2024.12.19.629451} uses diffusion models to predict cellular morphological responses to chemical and genetic perturbations. By adapting the Stable Diffusion framework with perturbation-specific encodings (scGPT for genetic data and RDKit for chemical compounds), MorphoDiff successfully generates high-resolution microscopy images that capture biologically meaningful phenotypic changes. However, MorphoDiff and similar methods focus on static images and therefore cannot capture the dynamic events—such as mitosis or migration—characteristic of time-lapse microscopy data. In contrast, our work extends diffusion-based generation into the temporal domain, adapting video diffusion models to the highly specialized setting of microscopy. This not only addresses the challenge of modeling subtle cellular behaviors over multiple frames, but also serves as a bridge between large-scale natural video models and domain-specific microscopy applications.

\section{Method}

\subsection{Dataset}

Our dataset comes from a large-scale RNAi-based phenotypic profiling study of HeLa cells under time-lapse microscopy \cite{Neumann2010}. In this screening campaign, each sample corresponds to a specific siRNA gene knockdown, targeting one of approximately 21,000 human genes. Each well was imaged every 30 minutes over a 48-hour period, generating time-lapse sequences that capture key events such as mitotic progression, migration, and cell death.

\paragraph{Phenotype Scoring.}
Each video in the screening is automatically assigned numeric phenotype scores along several axes, including:
\begin{itemize}
  \item cell death (frequency of dying cells),
  \item migration speed (average displacement of nuclei),
  \item proliferation (increased cell counts or divisions),
  \item mitotic delay, multi-lobed nuclei, ``grape-like'' morphology, etc.
\end{itemize}
For this work, we focus on four scores that are especially visually evident in time-lapse data: initial cell count, cell death, migration speed, and proliferation.

\paragraph{Filtering for Extreme Phenotypes.}
Although the original resource spans hundreds of plates and tens of thousands of wells, many knockdowns show only mild or clearly visibly discernible changes. We therefore apply percentile-based thresholding to the cell death, migration speed, and proliferation scores, labeling each well as \emph{HIGH}, \emph{LOW}, or \emph{MED} per phenotype (below the 10th percentile as LOW, above the 90th as HIGH, and in between as MED). We then retain only those wells that are extreme (HIGH or LOW) in at least two of the three phenotypes, ensuring clear visual contrasts. Finally, we truncate each time-lapse to the first 81 frames, discarding later frames to enforce consistent sequence lengths across all wells. After filtering, we obtain a final subset of about 3,500 videos at \(768 \times 1360\) resolution.

\subsection{Prompt Engineering Approach}
\label{sec:prompt_creation}

To guide our text-to-video generation, we convert the four numeric phenotype scores (initial cell count, proliferation, cell death, and migration speed) into concise textual prompts that describe visually observable cell behaviors. Specifically, each video is labeled as \emph{HIGH}, \emph{LOW}, or \emph{MED} for each phenotype, based on the percentile thresholds described earlier. We then translate these labels into short phrases:

\begin{itemize}
    \item \textbf{LOW initial cell count} $\rightarrow$ ``a few cells''
    \item \textbf{HIGH proliferation} $\rightarrow$ ``cells undergo frequent divisions''
    \item \textbf{HIGH migration} $\rightarrow$ ``cells move rapidly across the field''
    \item \textbf{HIGH cell death} $\rightarrow$ ``cells disappear frequently''
\end{itemize}
This mapping yields an unambiguous caption for each sample. 
For example, if a video is labeled \texttt{LOW initial cell count}, \texttt{LOW proliferation}, 
\texttt{HIGH migration}, and \texttt{HIGH cell death}, we create the visual prompt:

\begin{quote}
\small
``Time-lapse microscopy video of a few cells. The cells rarely divide, 
move rapidly, and frequently disappear due to cell death.''
\end{quote}
By avoiding extra details, these short textual prompts focus on the observable behaviors indicated by the phenotype scores, hopefully making them straightforward for the model to interpret. In other words, we aim to leverage the semantic priors of the pretrained language model, which can more readily interpret terms like ‘frequent divisions’ or ‘move rapidly’ than raw numeric values.

\subsection{Phenotype Embedding Approach}
\label{sec:phenotype_embedding}
As an alternative conditioning method, we directly embed numeric phenotype vectors into the model's latent space, removing reliance on textual descriptions. This numeric embedding approach potentially provides a more precise and direct way to inform the model about desired cellular behaviors. Specifically, we employ a single-token embedding:

\begin{itemize}
    \item We take a 4-dimensional phenotype vector 
    \[
      \mathbf{p} = 
      \bigl[
        p_{\text{cell\_count}},\,
        p_{\text{proliferation}},\,
        p_{\text{migration}},\,
        p_{\text{death}}
      \bigr]
      \in \mathbb{R}^4,
    \]
    where each entry is normalized to \([0,1]\).

    \item An MLP transforms \(\mathbf{p}\) into a 4096-dimensional embedding \(\mathbf{p}_{\text{embed}}\):
    \begin{align}
        \mathbf{h}_1 
          &= \mathrm{GELU}(\mathbf{W}_1 \,\mathbf{p} + \mathbf{b}_1) 
             \in \mathbb{R}^{256},\\
        \mathbf{h}_2 
          &= \mathrm{GELU}(\mathbf{W}_2 \,\mathbf{h}_1 + \mathbf{b}_2)
             \in \mathbb{R}^{512},\\
        \mathbf{p}_{\mathrm{embed}} 
          &= \mathbf{W}_3\,\mathbf{h}_2 + \mathbf{b}_3
             \in \mathbb{R}^{4096}.
    \end{align}

    **4096 was chosen to align with the token embedding dimension used by the text encoder, thereby ensuring the phenotype embedding can be treated as a single token within the same latent space.**

    \item We prepend \(\mathbf{p}_{\mathrm{embed}}\) as a single additional token to the text token embeddings 
    \(\mathbf{E}_{\mathrm{text}} \in \mathbb{R}^{L \times d_{\mathrm{model}}}\):
    \begin{align}
      \mathbf{E}_{\mathrm{final}}
      \;=\;
       \bigl[\;\mathbf{p}_{\mathrm{embed}};\,
             \mathbf{E}_{\mathrm{text}}\bigr]
       \;\in\;\mathbb{R}^{(L+1)\times 4096}.
    \end{align}
\end{itemize}
This single-token scheme consolidates all phenotype information into one 4096-dimensional vector, which the transformer can attend to alongside the text tokens (e.g., a minimal prompt like \emph{“Time-lapse microscopy video of cells”}). The model is fine-tuned end-to-end, learning to map these numeric values to appropriate visual representations in the generated videos. For the embedding approach, we train the CogVideoX backbone and the phenotype embedding MLP jointly. At inference time, setting the phenotype vector \(\mathbf{p}\) enables direct numeric control of the generated video—\emph{e.g.}, raising \(p_{\text{proliferation}}\) should increase the frequency of cell division in the final animation.

\subsection{Training Configurations}
\label{sec:training_configs}

We selected CogVideoX as our base model among existing open-source video diffusion models for two primary reasons: (1) it achieves state-of-the-art performance on generating time-lapse videos~\cite{yuan2024chronomagicbenchbenchmarkmetamorphicevaluation}, and (2) it provides pretrained checkpoints for both text-to-video (T2V) and image-to-video (I2V) generation, unlike alternative models such as HunyuanVideo~\cite{kong2025hunyuanvideosystematicframeworklarge}, which currently support only text-conditioned generation. Having access to both T2V and I2V checkpoints allowed us to comprehensively explore multiple conditioning strategies. To evaluate the impact of phenotype conditioning, we trained and compared four distinct model configurations using the CogVideoX backbone:

\paragraph{(1) Unconditional Baseline.}
We omit all phenotype signals and provide the model with only a minimal, generic prompt (e.g., ``Time-lapse microscopy video of cells''). This model learns to generate cell sequences without explicit control over initial cell count, proliferation, migration, or cell death.

\paragraph{(2) Text-Prompted Conditioning.}
Numeric phenotype scores are translated into concise textual captions (e.g., \emph{``cells rarely divide, move rapidly, and frequently disappear due to cell death''}), guiding the model to explicitly represent specific cellular behaviors during generation.

\paragraph{(3) Numeric Phenotype Embedding.}
Numeric phenotype scores are embedded into a single 4096-dimensional latent token through an MLP and prepended to minimal textual prompts. The model thus directly interprets numeric inputs for phenotype-conditioned generation, removing reliance on descriptive text.

\paragraph{(4) Image-conditioned Generation (Image2Video).}
In addition to text-based conditioning, we explore conditioning the model directly on an initial real microscopy frame. Specifically, we first encode the real microscopy image with the same 3D VAE used by the videos, then concatenate its latent representation to the noised video input along the channel dimension at each diffusion step. This approach is inherently valuable because it allows researchers to specify initial cellular states explicitly, enabling the direct visualization of how individual cells or cell clusters evolve over time. Such fine-grained control can be particularly useful for simulating specific experimental conditions or for understanding dynamic cellular behaviors initiated from observed biological states.

\vspace{1em}
\noindent\textbf{Implementation Details.}  
All four configurations were trained on the same curated subset of approximately 3,500 microscopy videos using identical hyperparameters: a global batch size of 4, a fixed learning rate, and the same total training steps. While we initially explored LoRA (low-rank adaptation) \cite{hu2021loralowrankadaptationlarge} across both text-to-video and image-to-video modes, preliminary results showed that LoRA underperformed full fine-tuning in text-to-video settings. We therefore only used a full fine-tune for text-based conditioning. In contrast, the stronger conditioning signal from a real microscopy frame allowed LoRA to achieve competitive results for the image-to-video model, where we trained both LoRA and full fine-tuned models. Each training run utilized 4 A100 GPUs (80 GB VRAM each), and DeepSpeed ZeRO-2 optimization was utilized to efficiently train on multiple GPUs.

\section{Evaluation}

We evaluate our models on both morphological accuracy and phenotype alignment. In particular, we focus on whether the generated cell shapes remain plausible at a per-frame level and whether each model accurately reproduces population dynamics over time.

\paragraph{Morphological Analysis.}
We segment each frame of both real and generated videos using nucleus-segmentation methods from Cellpose \cite{Stringer2020CellposeAG}. For each nucleus, we measure four shape descriptors that capture the following morphological properties:
\emph{Area}, \emph{Eccentricity}, \emph{Solidity}, and \emph{Perimeter}.

\paragraph{Population Analysis}
To assess proliferation and cell death, we track the total number of nuclei in each frame throughout the video sequence. We measure the \emph{initial cell count} (cells in the first frame), \emph{final cell count}, and the \emph{growth ratio} (final count divided by initial count). Additionally, we estimate the number of \emph{division events} by counting the appearances of new nuclei.

\paragraph{Movement Analysis.}
To quantify cell migration, we track the centroid positions of individual nuclei throughout the sequence. We measure the \emph{net displacement}, defined as the straight-line distance between each nucleus’s initial and final positions, as well as the \emph{average migration speed}. Higher net displacement values correspond to cells migrating greater overall distances, aligning with conditions labeled as “HIGH migration.”

\begin{figure}[ht]
    \centering
    \begin{subfigure}[b]{0.46\textwidth}
        \includegraphics[width=\textwidth]{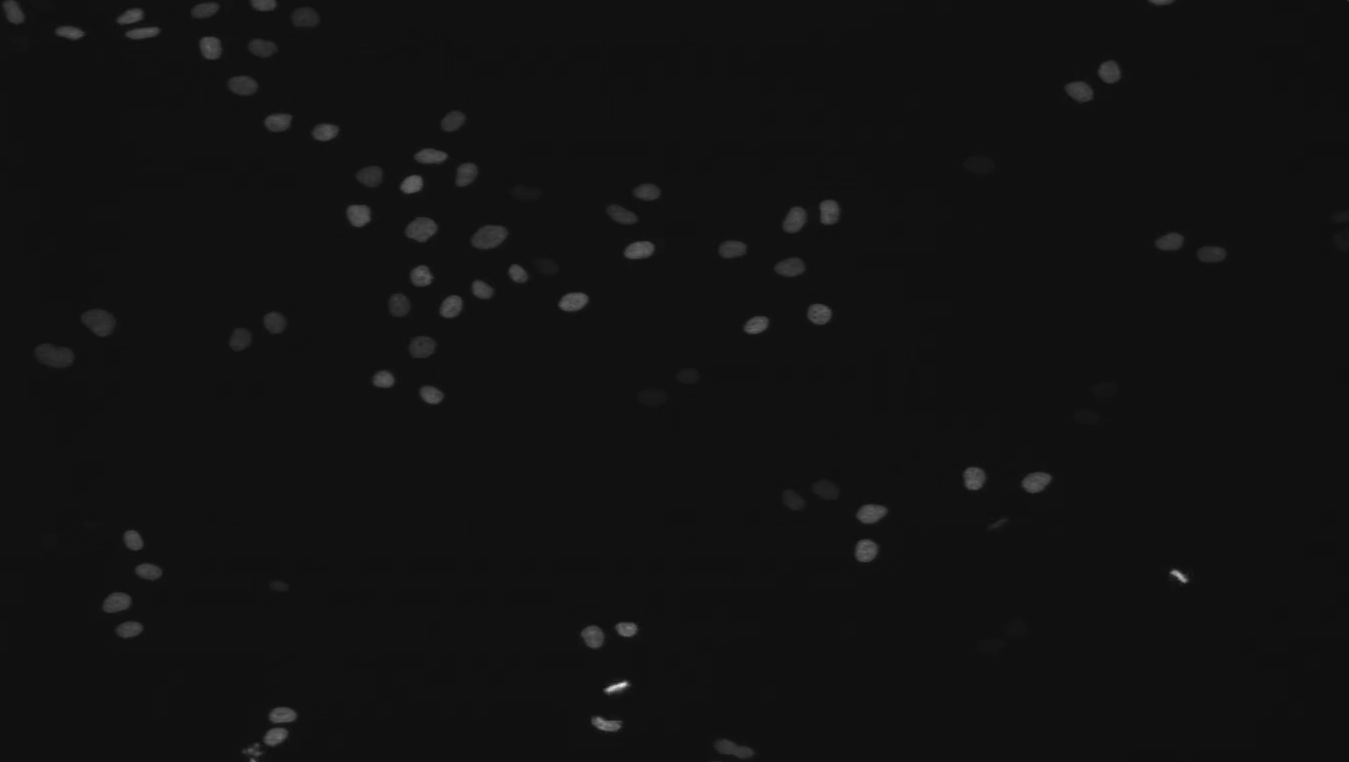}
        \caption{Raw frame from the microscopy video.}
        \label{fig:raw_frame}
    \end{subfigure}
    \hfill
    \begin{subfigure}[b]{0.46\textwidth}
        \includegraphics[width=\textwidth]{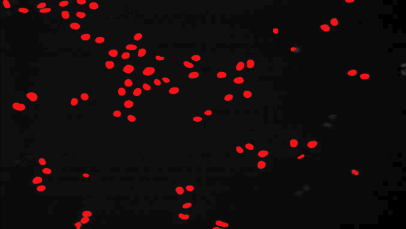}
        \caption{Segmentation overlay on the same frame.}
        \label{fig:seg_overlay}
    \end{subfigure}
    \caption{%
    Example nucleus segmentation on a single frame. We apply this segmentation 
    pipeline to \emph{every} frame in each time-lapse, extracting per-nucleus shape 
    descriptors (e.g., area, eccentricity, solidity, perimeter). These values are then 
    aggregated across frames to compute the population-level morphological metrics 
    used in our evaluation.
    }
    \label{fig:nucleus_segmentation_example}
\end{figure}

\paragraph{Phenotype Alignment.}
Because each well is labeled (or numerically encoded) with specific phenotypes (e.g., HIGH vs.\ LOW proliferation), we verify that generated outcomes reflect these labels. For instance, a model conditioned on \emph{HIGH} proliferation should produce a significantly larger final cell count distribution than one conditioned on \emph{LOW}. 

\paragraph{Comparative Analysis Across Model Variants.}
We compare the four training configurations (unconditional, text-conditioned, numeric-conditioned, and image-conditioned) by evaluating how well each model reproduces the phenotype distributions found in the test set. Specifically, for each phenotype category (e.g., \emph{HIGH} proliferation), we gather all real test-set videos labeled with that category and generate a corresponding batch of synthetic videos, each conditioned (via text prompts or numeric embeddings) to the same phenotype. We then compute morphological, population, and movement metrics on both real and generated videos and measure the discrepancy of their distributions using the one-dimensional \emph{Wasserstein-1} distance:
\begin{equation}    
W_1(F,G) \;=\; \int_0^1 \Bigl| F^{-1}(u)\;-\;G^{-1}(u) \Bigr|\;\mathrm{d}u
\end{equation}
where \(F\) and \(G\) denote the real and synthetic distributions, respectively. A lower \(W_1\) indicates higher similarity. By repeating this procedure for each label (\emph{HIGH} vs. \emph{LOW}) and each phenotype (proliferation, migration speed, etc.), we assess whether the generated videos exhibit appropriately distinct behaviors aligned with the specified phenotypic conditions.

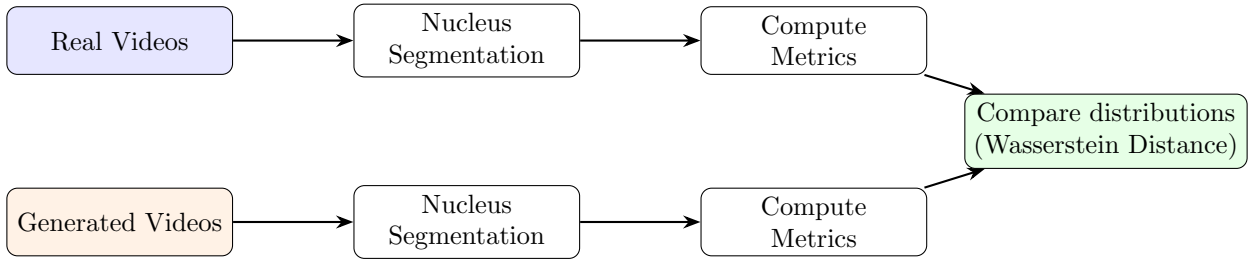
\begin{figure}[ht]
    \centering
    \begin{tikzpicture}[
        node distance=1.6cm,
        box/.style={rectangle, draw, rounded corners, minimum width=3cm, minimum height=0.9cm, align=center},
        arrow/.style={-{Stealth}, thick},
        label/.style={font=\small}
    ]

    \node[box, fill=blue!10] (realvid) {Real Videos};
    \node[box, right=of realvid] (realseg) {Nucleus\\Segmentation};
    \node[box, right=of realseg] (realmetric) {Compute\\Metrics};

    \node[box, fill=orange!10, below=1.5cm of realvid] (genvid) {Generated Videos};
    \node[box, right=of genvid] (genseg) {Nucleus\\Segmentation};
    \node[box, right=of genseg] (genmetric) {Compute\\Metrics};

    \node[box, fill=green!10, right=2cm of $(realmetric)!0.5!(genmetric)$] (compare) {Compare distributions\\(Wasserstein Distance)};

    \draw[arrow] (realvid) -- (realseg);
    \draw[arrow] (realseg) -- (realmetric);

    \draw[arrow] (genvid) -- (genseg);
    \draw[arrow] (genseg) -- (genmetric);

    \draw[arrow] (realmetric) -- (compare);
    \draw[arrow] (genmetric) -- (compare);

    \end{tikzpicture}

    \caption{Overview of evaluation pipeline. Metrics computed from real and generated videos are compared using Wasserstein distance to assess realism and phenotype alignment.}
    \label{fig:evaluation_pipeline}
\end{figure}

\section{Experiments and Results}

\begin{figure}[H]
\centering
\begin{tabular}{c@{\hspace{3mm}}c@{\hspace{2mm}}c@{\hspace{2mm}}c@{\hspace{2mm}}c}
 & Frame 0 & Frame 20 & Frame 60 & Frame 80 \\[2mm]
\raisebox{4ex}{\textbf{CogVideoX}} &
\includegraphics[width=0.20\textwidth]{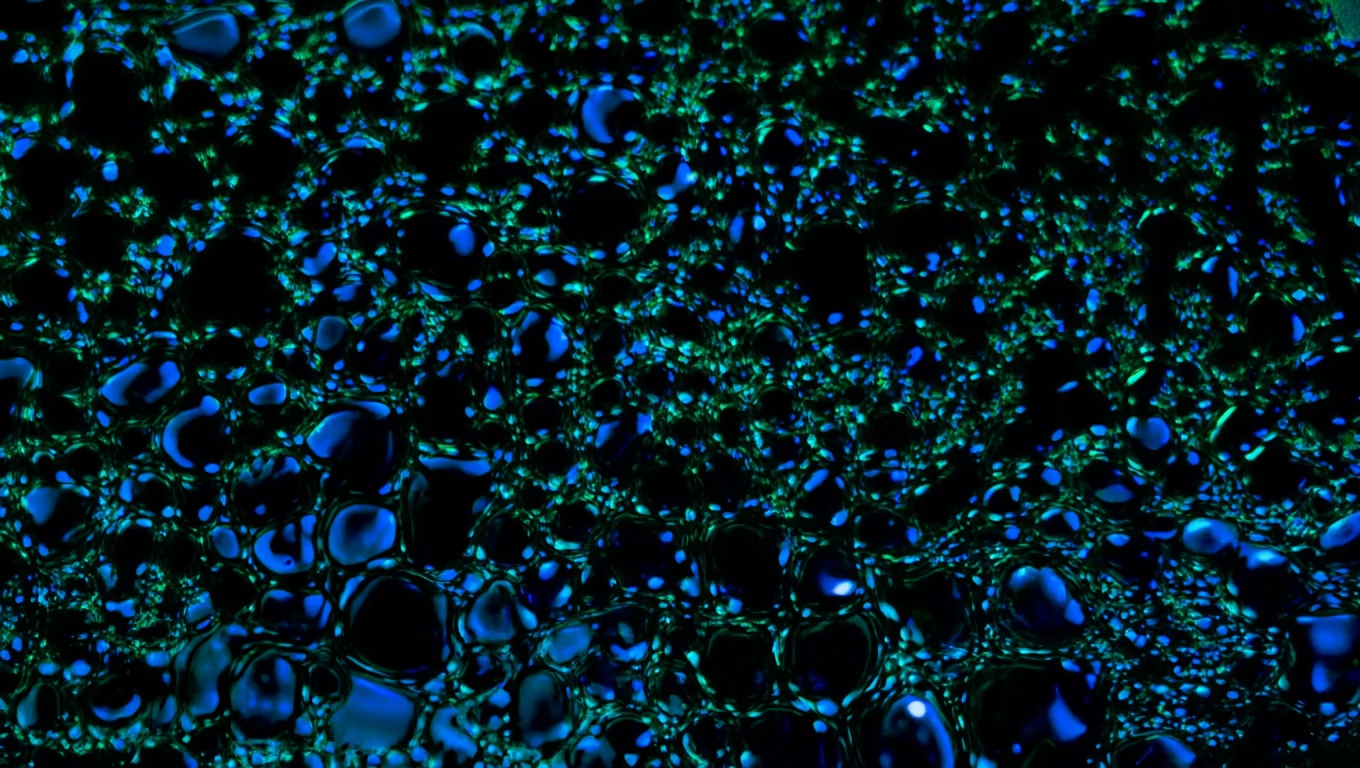} &
\includegraphics[width=0.20\textwidth]{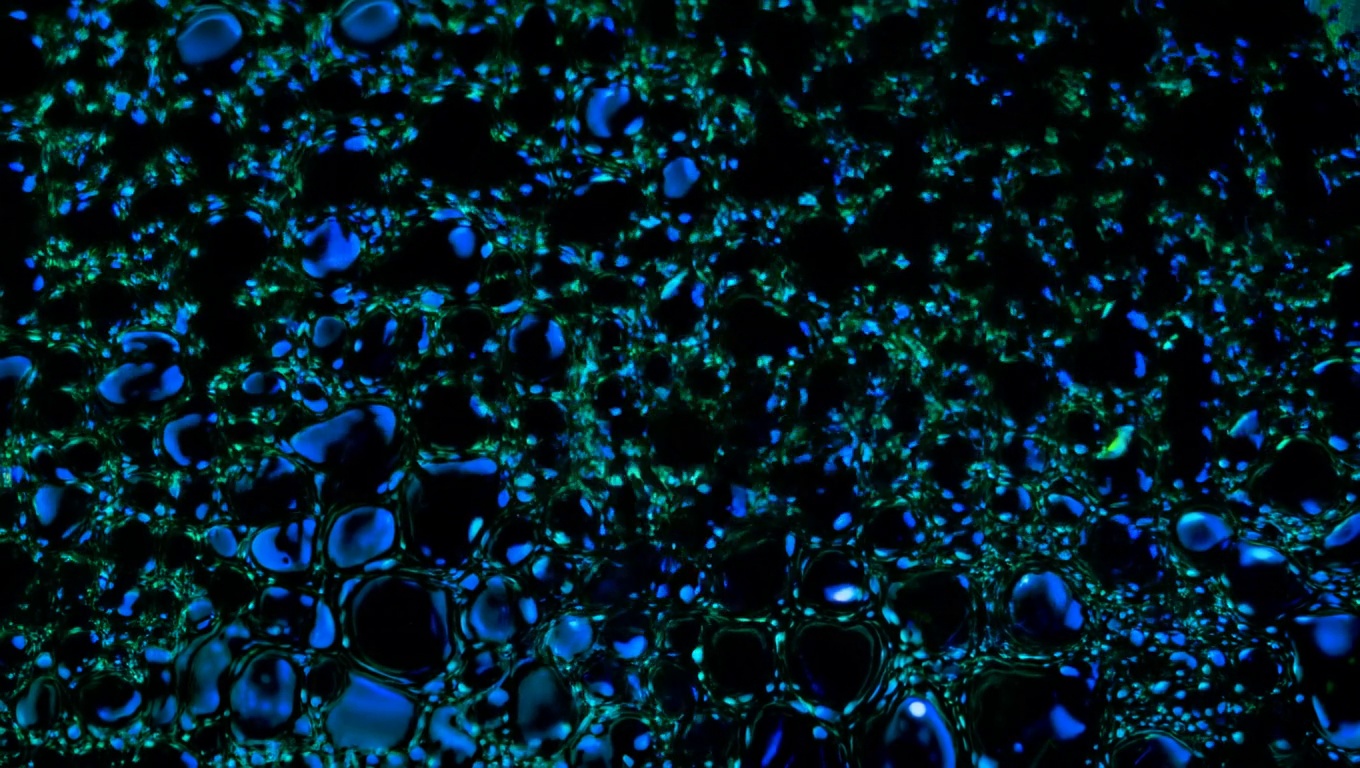} &
\includegraphics[width=0.20\textwidth]{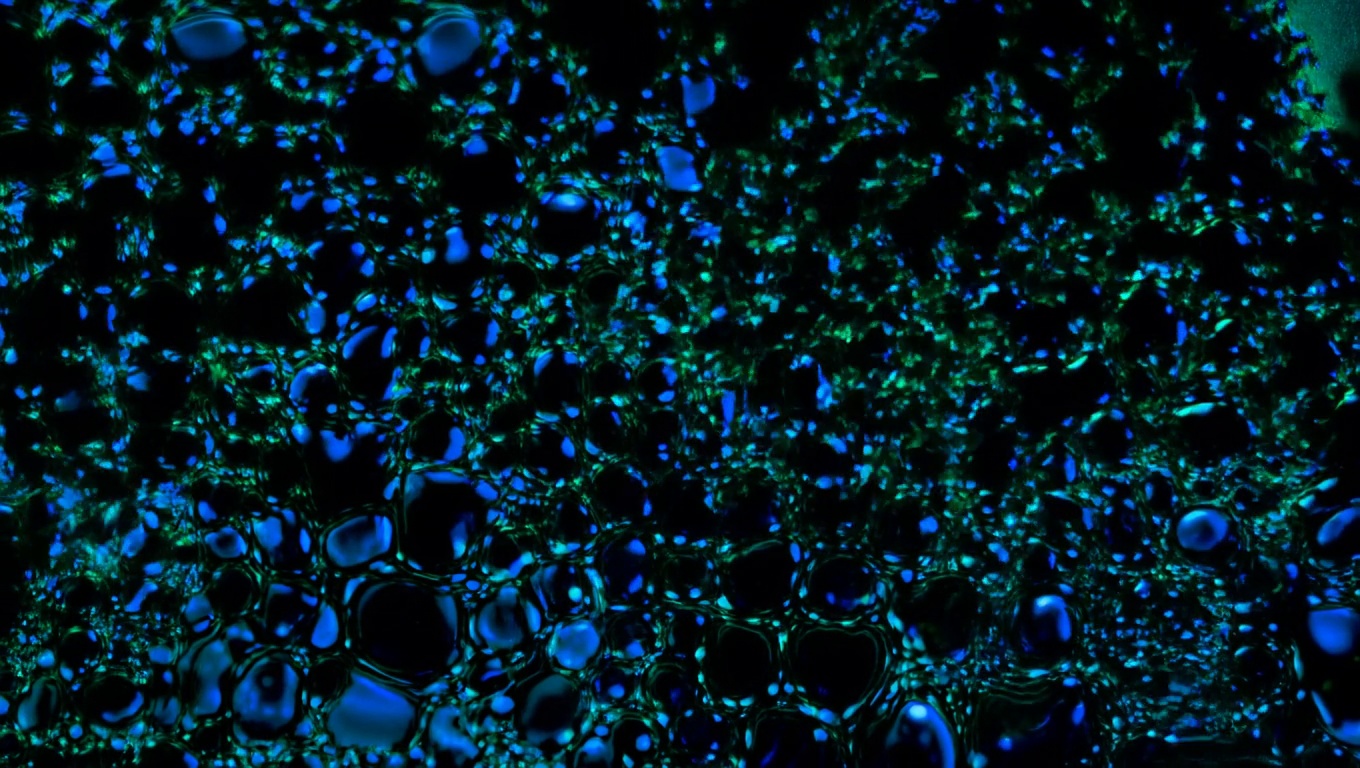} &
\includegraphics[width=0.20\textwidth]{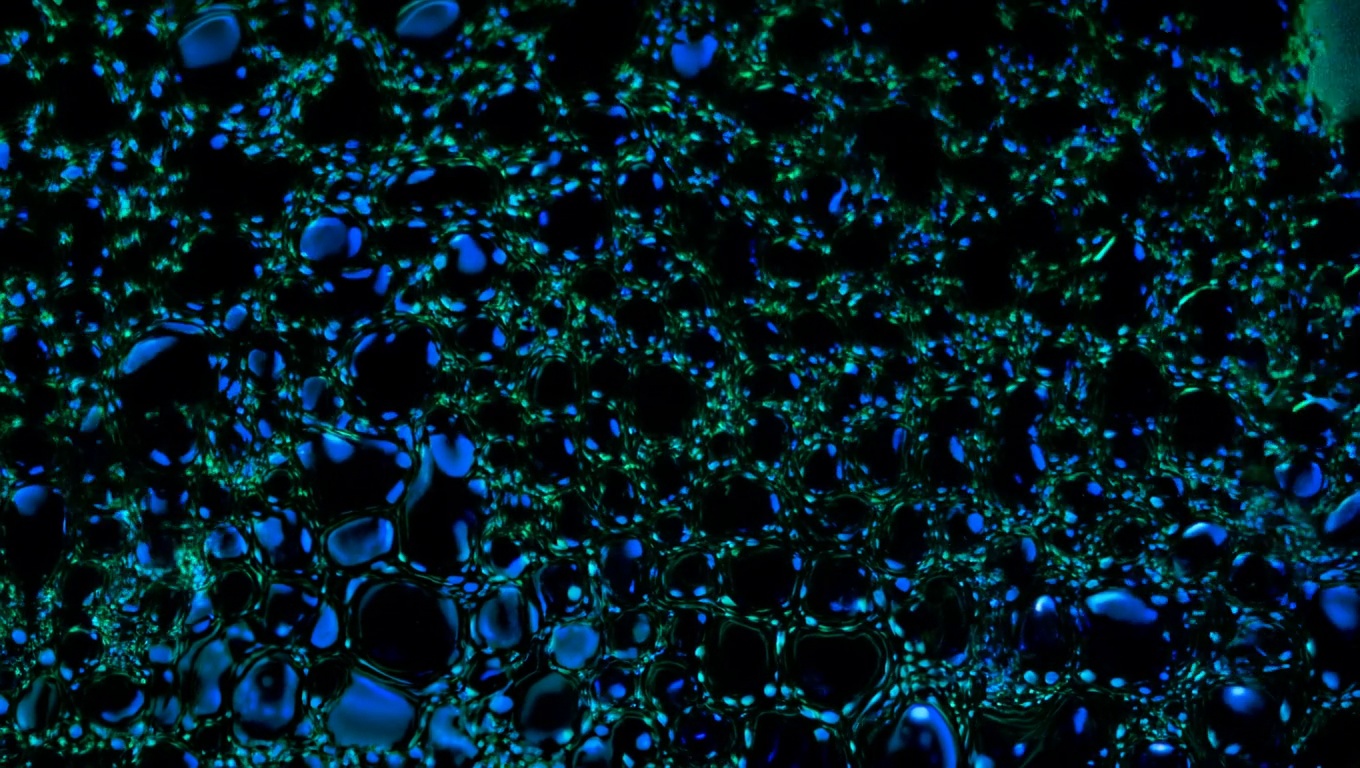} \\[1mm]
\raisebox{4ex}{\textbf{Fine-tuned}} &
\includegraphics[width=0.20\textwidth]{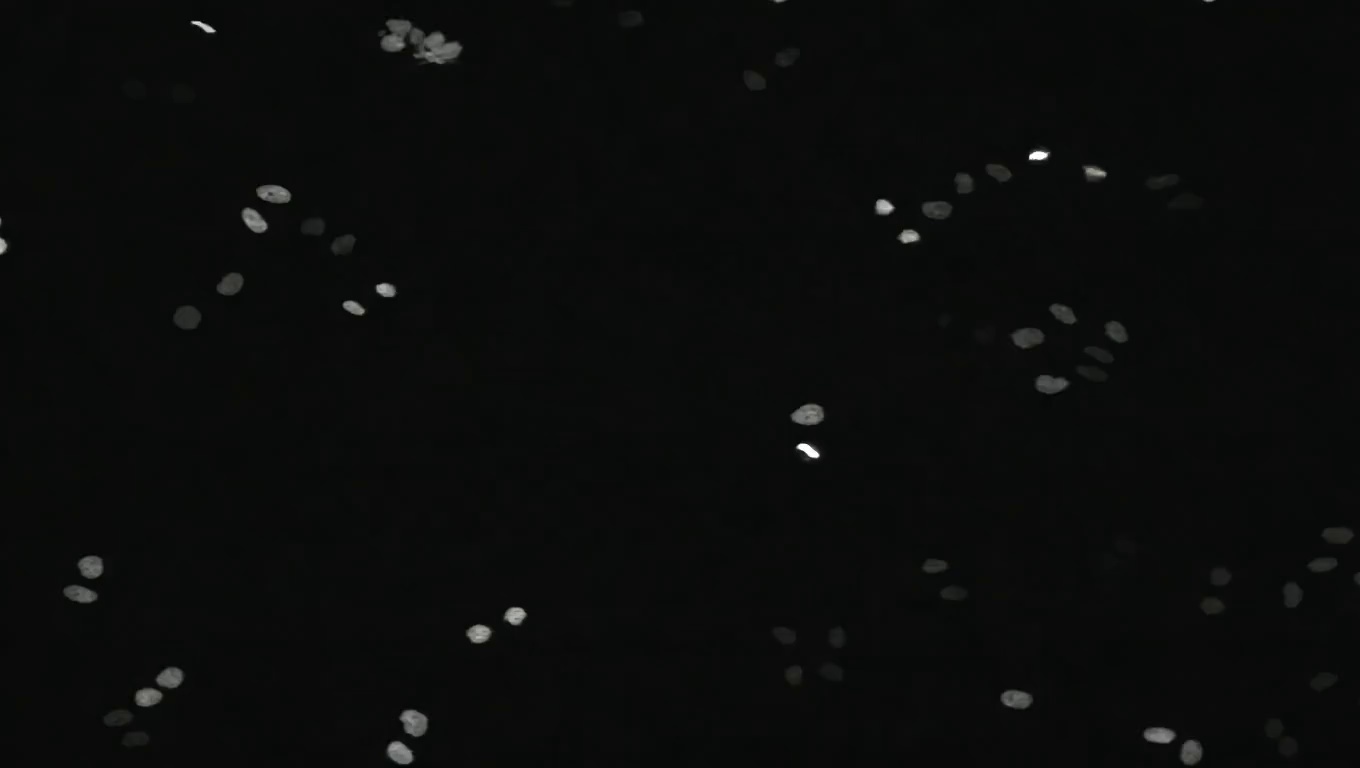} &
\includegraphics[width=0.20\textwidth]{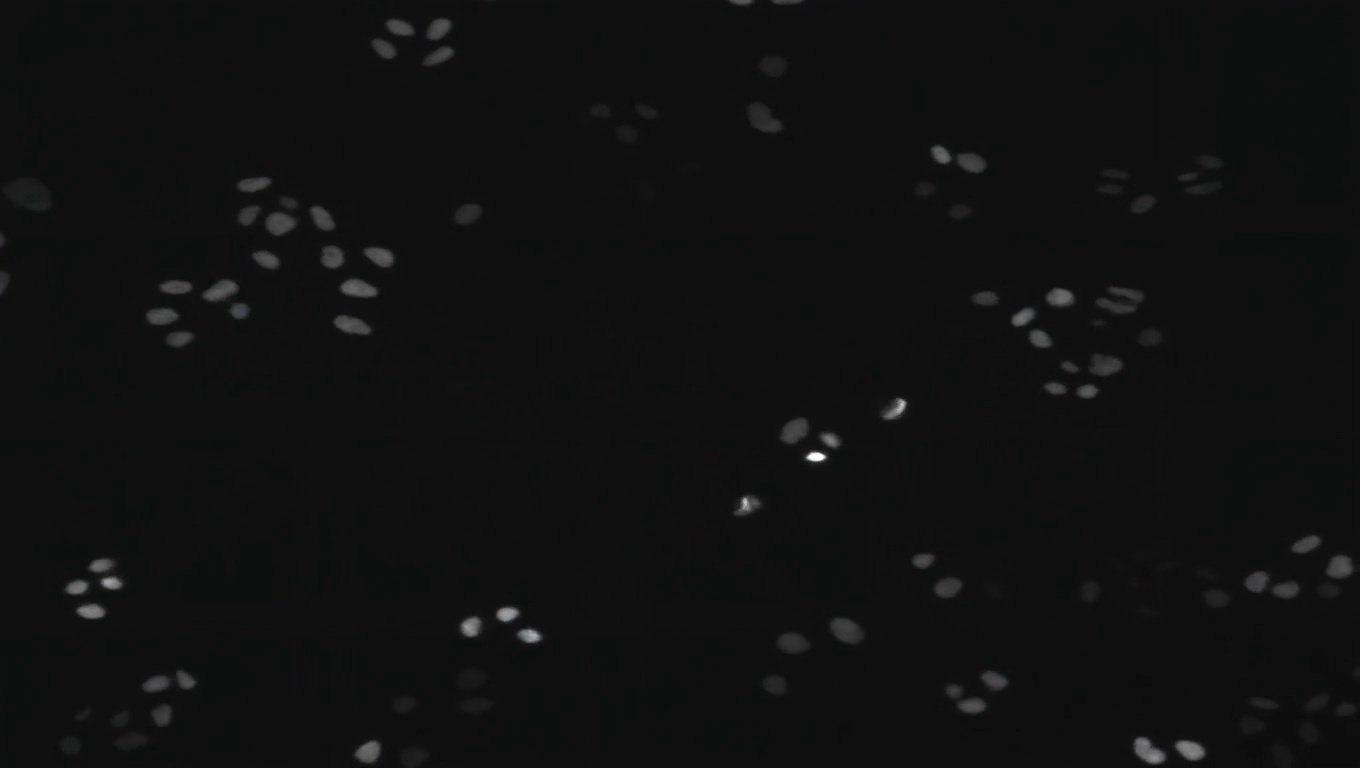} &
\includegraphics[width=0.20\textwidth]{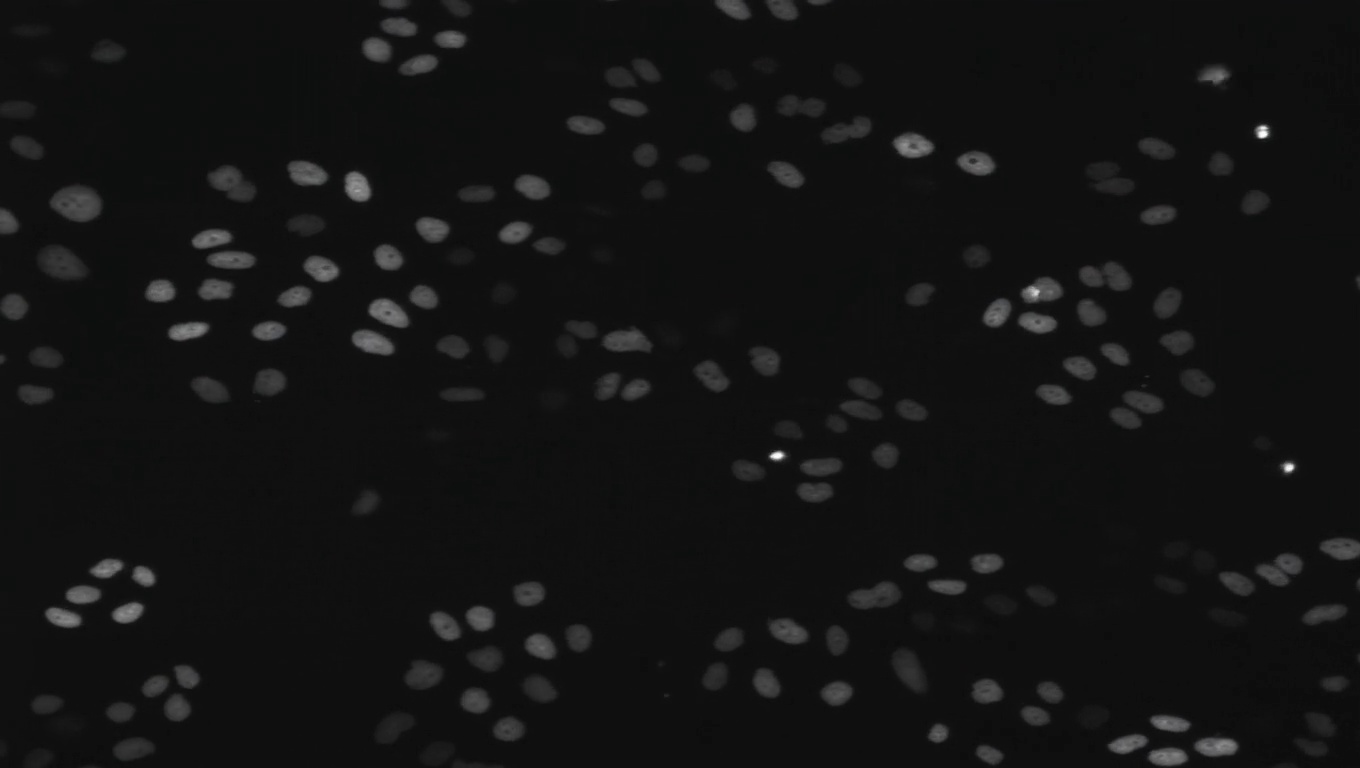} &
\includegraphics[width=0.20\textwidth]{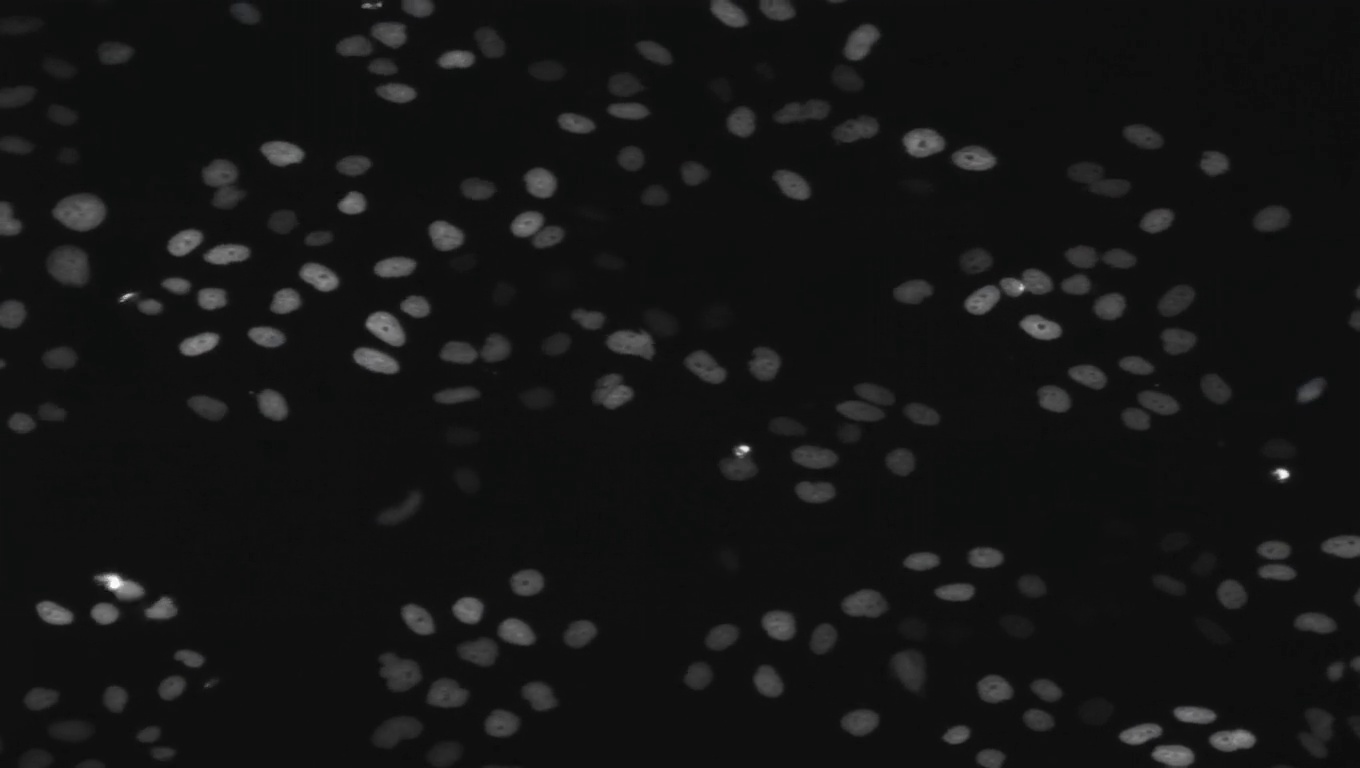} \\[1mm]
\raisebox{4ex}{\textbf{Real}} &
\includegraphics[width=0.20\textwidth]{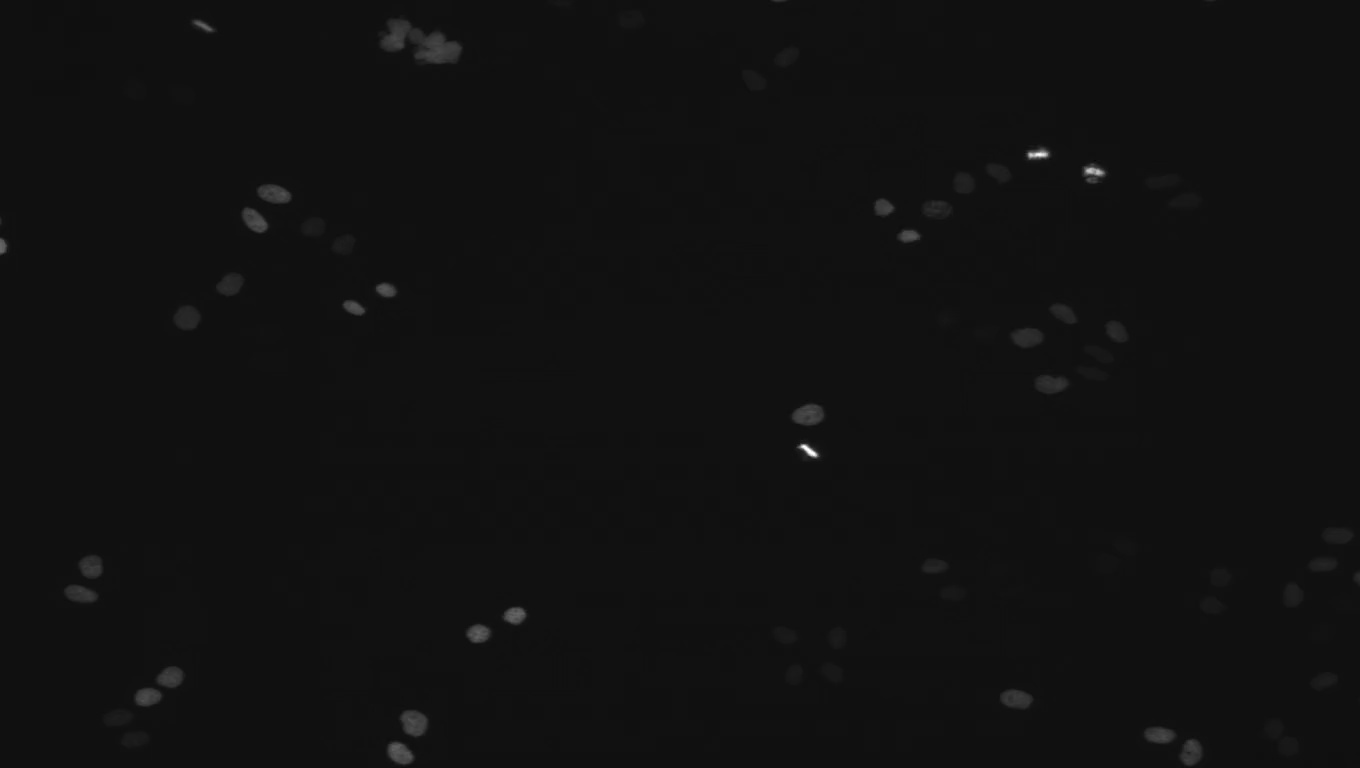} &
\includegraphics[width=0.20\textwidth]{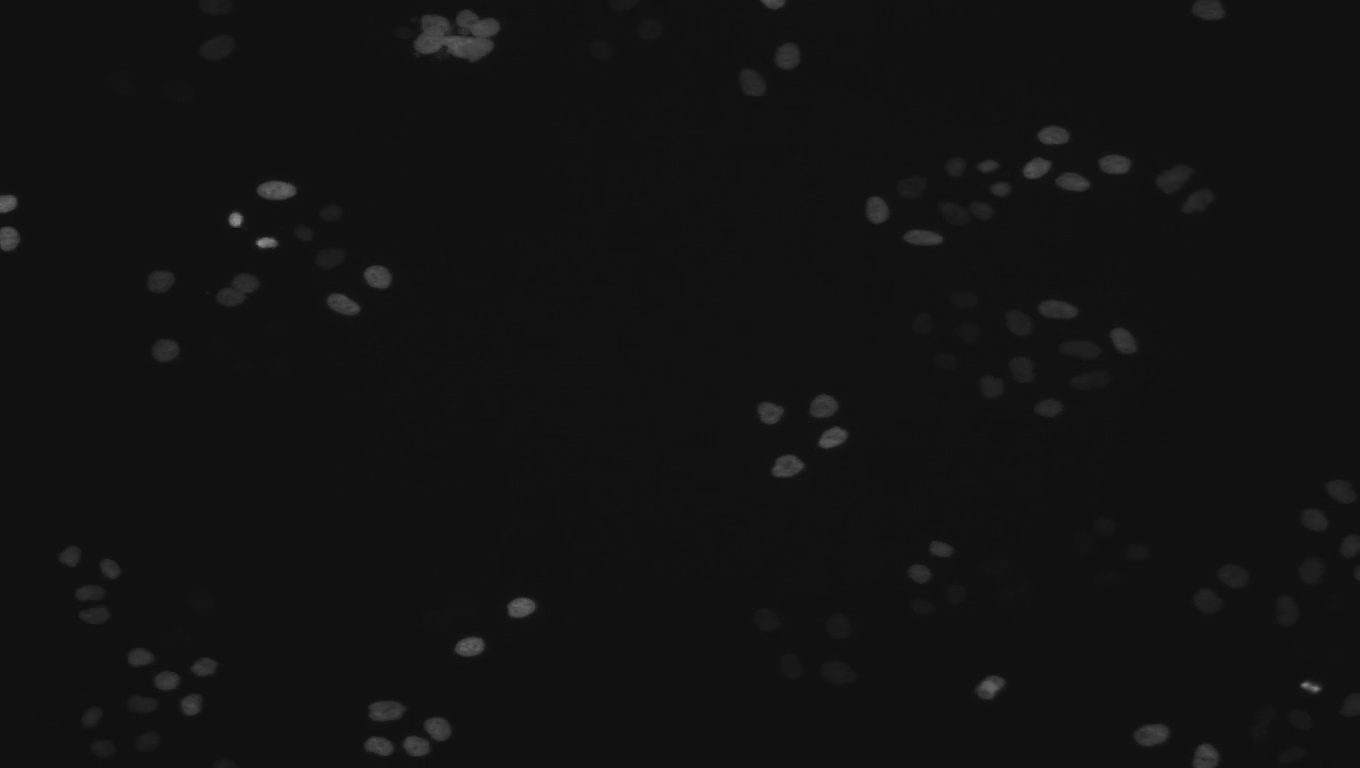} &
\includegraphics[width=0.20\textwidth]{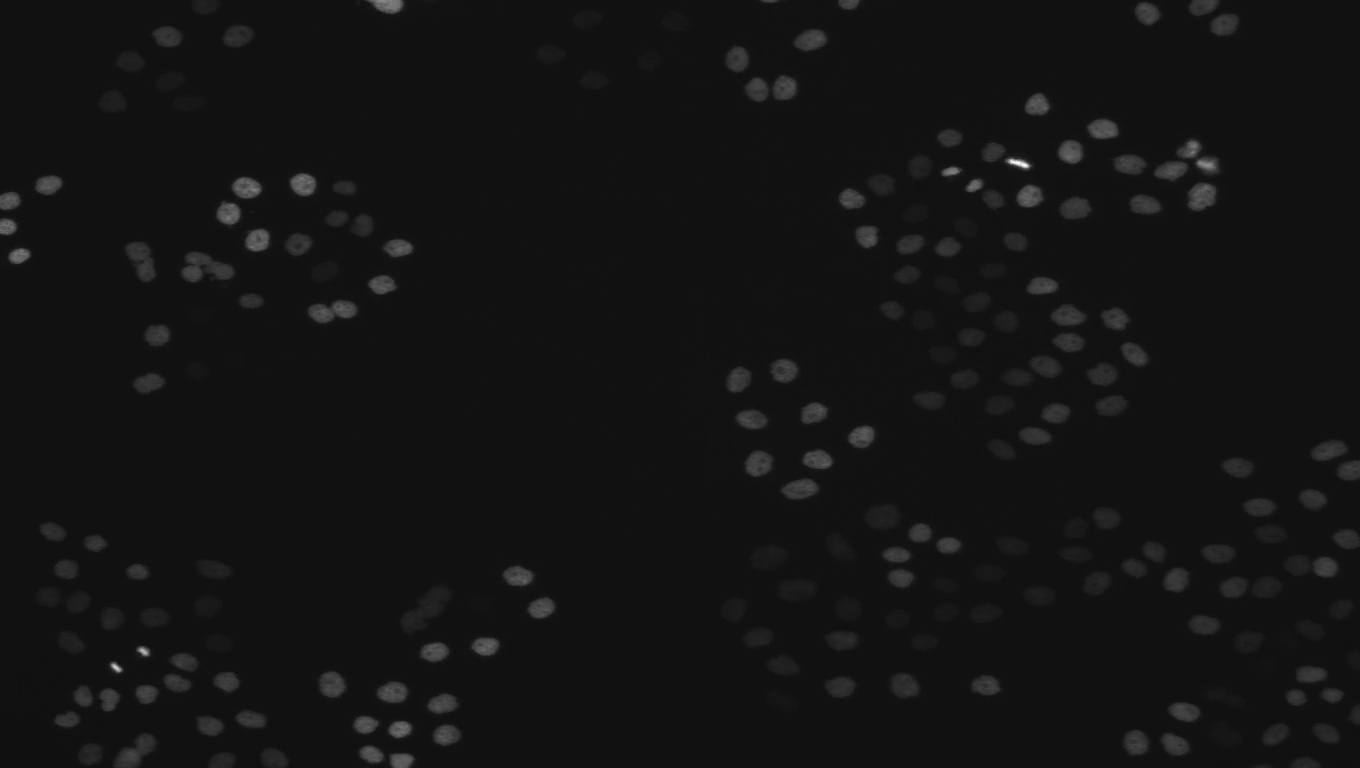} &
\includegraphics[width=0.20\textwidth]{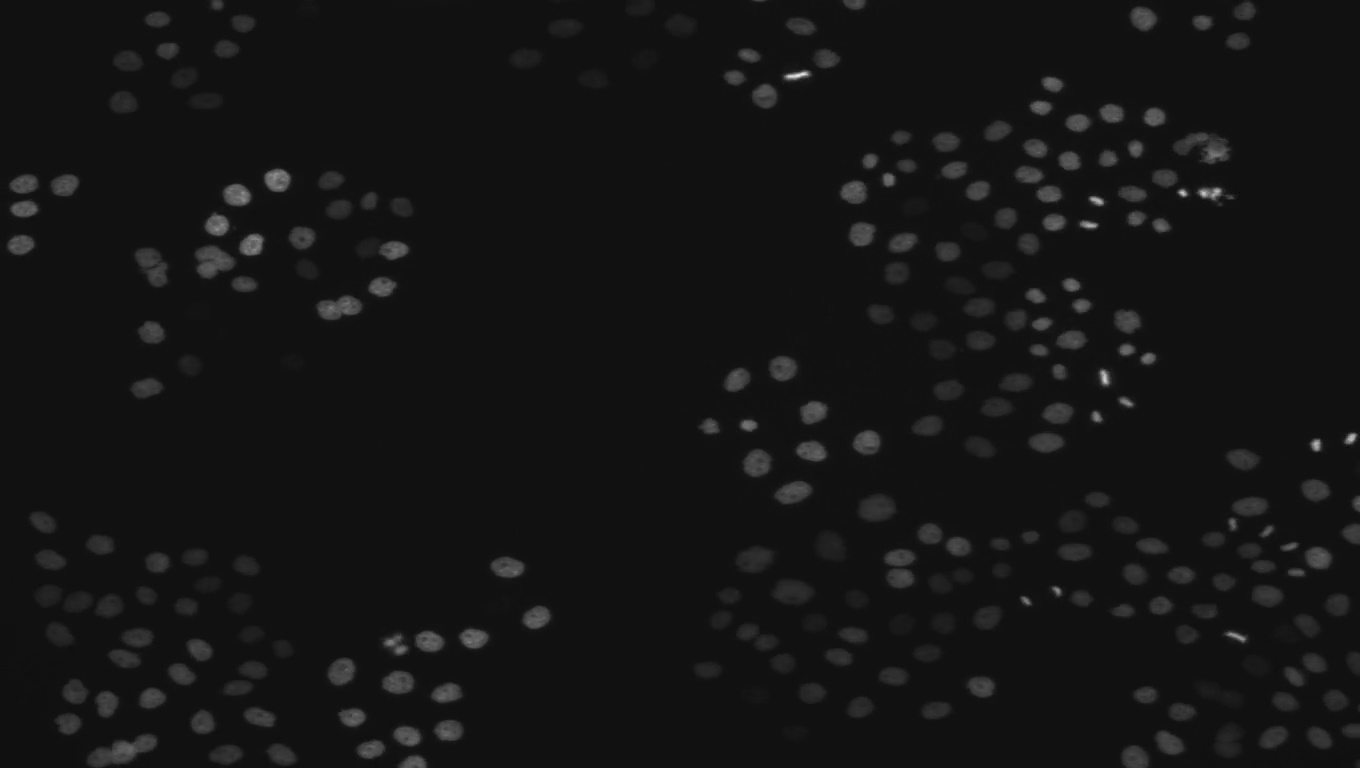} \\
\end{tabular}
\caption{
Visual comparison of zero-shot CogVideoX (top row), fine-tuned CogVideoX (middle row), 
and real microscopy frames (bottom row) at four time points (0, 20, 60, 80). 
The zero-shot baseline produces unrealistic backgrounds and fails to capture cell divisions, 
whereas the fine-tuned model generates biologically plausible mitotic events and 
cell population growth patterns that closely resemble real data. 
}
\label{fig:vis_samples1}
\end{figure}

\subsubsection{Morphology Metrics}
Table~\ref{tab:morphology_results} presents actual morphological descriptor values (\emph{Area}, \emph{Eccentricity}, \emph{Solidity}, and \emph{Perimeter}) for nuclei segmented from real and generated microscopy sequences. We compare the unconditional fine-tuned model with the phenotype-conditioned models (text-conditioned and numeric-conditioned). The zero-shot baseline is excluded from this analysis because the pretrained CogVideoX model, not fine-tuned on microscopy data, produces videos fundamentally different from our microscopy sequences (see Figure~\ref{fig:vis_samples1}), making biological metrics meaningless in that context. Since the phenotype conditioning primarily targets population dynamics and migration behaviors, we do not anticipate substantial differences in nucleus-level morphology between these model variants and we therefore present direct values (mean ± SD) for the morphological metrics rather than Wasserstein distances to the real distribution. 

\begin{table}[H]
\centering
\caption{Morphology Metrics: Mean and standard deviation (Mean ± SD) of morphological descriptors for nuclei in real and generated videos.}
\vspace{1em}
\label{tab:morphology_results}
\begin{tabular}{lcccc}
    \toprule
    \textbf{Condition} & \textbf{Area ($pixel^2$)} & \textbf{Eccentricity} & \textbf{Solidity} & \textbf{Perimeter ($pixel$)} \\
    \midrule
    Real Videos                    & 412.6 ± 165 & 0.711 ± 0.141 & 0.958 ± 0.017 & 75.5 ± 17.1 \\
    Unconditional (FT)    & 367.4 ± 157 & 0.718 ± 0.140 & 0.956 ± 0.016 & 71.4 ± 17.1 \\
    Text-Conditioned (FT) & 452.5 ± 204 & 0.712 ± 0.148 & 0.956 ± 0.023 & 78.8 ± 22.1 \\
    Numeric-Conditioned (FT) & 379.2 ± 250 & 0.700 ± 0.169 & 0.950 ± 0.031 & 69.2 ± 29.0 \\
    \bottomrule
\end{tabular}
\end{table}

\subsubsection{Population and Movement Metrics}
Table~\ref{tab:population_results} compares models using metrics related to population dynamics and migration speed under HIGH and LOW conditions. Fine-tuned models (unconditional, text-conditioned, numeric-conditioned) substantially outperform the zero-shot baseline, accurately reflecting realistic dynamics such as cell division and migration. However, the explicit conditioning methods (text and numeric) fail to produce clear differences between HIGH and LOW conditions, suggesting that phenotype scores alone were insufficient to exert strong control over the population dynamics.

\begin{table}[H]
    \centering
    \caption{Population and Movement Metrics: Wasserstein distances for real vs.\ generated data under HIGH and LOW phenotype conditions (lower is better).}
    \vspace{1em}
    \label{tab:population_results}
    \begin{tabular}{lcc|cc|cc}
        \toprule
        & \multicolumn{2}{c}{\textbf{Final Cell Count}} & \multicolumn{2}{c}{\textbf{Growth Ratio}} & \multicolumn{2}{c}{\textbf{Net Displacement}} \\
        \textbf{Model} & \textbf{HIGH} & \textbf{LOW} & \textbf{HIGH} & \textbf{LOW} & \textbf{HIGH} & \textbf{LOW} \\
        \midrule
        Unconditional (Fine-tuned) & \textbf{80.47} & 295.92 & \textbf{0.47} & 1.34 & \textbf{5.92} & \textbf{0.70} \\
        Text-Conditioned (Fine-tuned) & 379.50 & 55.04 & 2.84 & \textbf{1.21} & 10.52 & 4.04 \\
        Numeric-Conditioned (Fine-tuned) & 343.18 & \textbf{17.06} & 2.31 & 1.27 & 9.14 & 2.65 \\
        \bottomrule
    \end{tabular}
\end{table}

Overall, fine-tuning our video diffusion model on microscopy data leads to more realistic sequences than the zero-shot baseline, as shown in Figure~\ref{fig:vis_samples1}. Incorporating phenotype information (through text or numeric embeddings) offers a direct way to specify desired cell behaviors, but it does not strongly differentiate subtle phenotypes in our current setup. The frames in Figure~\ref{fig:vis_samples1} highlight that fine-tuned models capture cell divisions and morphological changes better than the zero-shot baseline, which exhibits minimal variation over time. This contrast becomes even clearer when viewing full video sequences: the fine-tuned model’s temporally coherent dynamics stand in stark relief against the baseline’s static or inconsistent outputs (see Supplementary Material for full videos).

\subsection{Extended Sequence Generation}
\label{sec:extended_sequence}

\begin{figure}[H]
  \centering
  \includegraphics[width=0.8\linewidth]{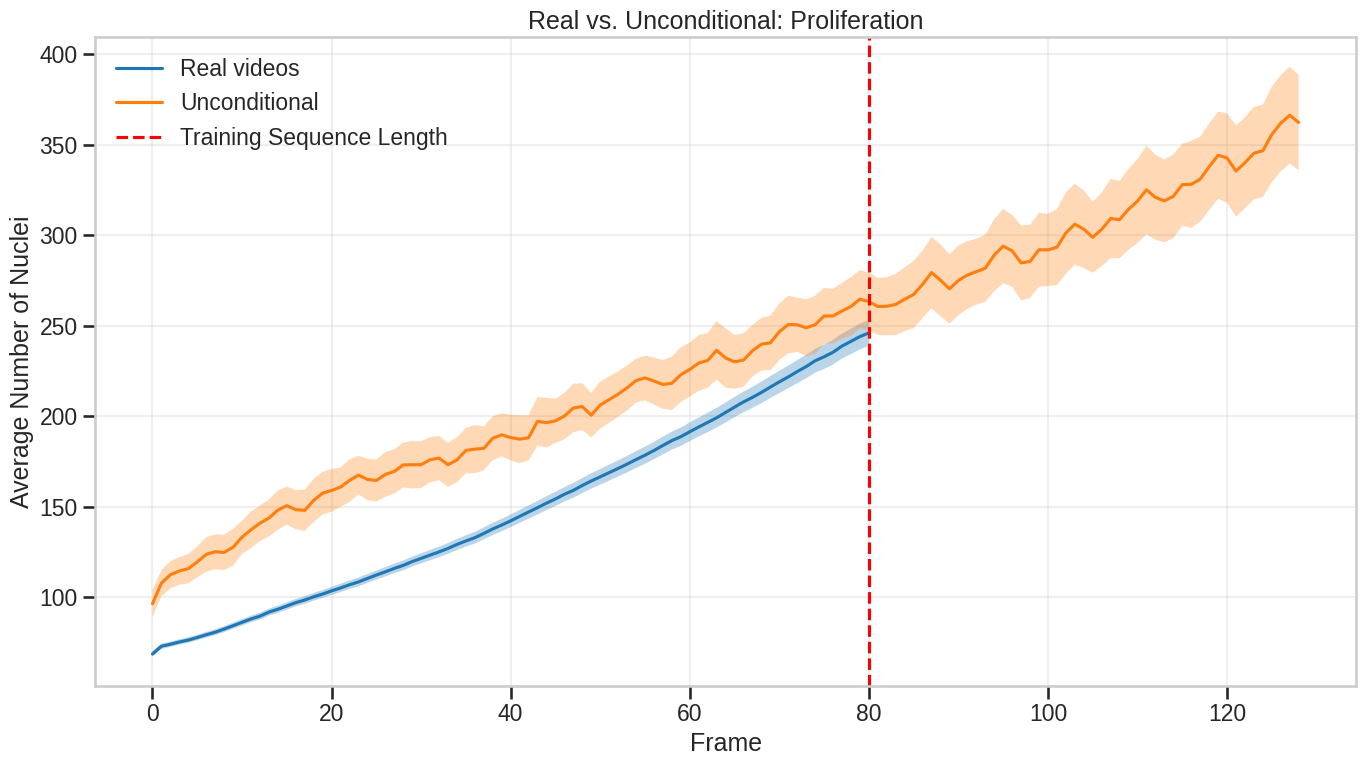} 
  \caption{
    Comparison of average cell counts between real videos and 
    our unconditional model, extended to 129 frames. The shaded region indicates the standard deviation across multiple sequences.
  }
  \label{fig:extended_sequence}
\end{figure}

Our model was trained on 81-frame sequences, but we tested its ability to generate longer time-lapses by extending outputs to 129 frames. To assess stability beyond the training horizon, we tracked cell counts from frames 0 to 129. As shown in Figure~\ref{fig:extended_sequence}, the fine-tuned model continues to produce realistic proliferation trends well past frame 81, closely matching real data. This result suggests the model has learned stable temporal patterns rather than simply memorizing fixed-length sequences, demonstrating its capacity to extrapolate biological behaviors over extended durations. This ability to extrapolate well beyond the training horizon could be particularly valuable for long-term biological studies, where lab protocols may require imaging cells for several days. By generating realistic cell trajectories at extended time scales, our approach can potentially support in silico experiments, for example allowing researchers to visualize outcomes of prolonged treatments or delayed phenotypes without physically capturing further real-world frames.

\subsection{Image-to-Video (I2V) Results}
\label{sec:i2v}

We now evaluate models conditioned on an initial microscopy frame rather than text or numeric prompts. In Table~\ref{tab:i2v_morphology}, we compare the zero-shot CogVideoX baseline, 
a LoRA (low-rank adaptation) variant, and a fully fine-tuned model in terms of morphological metrics. While the zero-shot baseline displays moderately low Wasserstein distances, it effectively “freezes” the initial frame, failing to capture subsequent cell divisions or morphological changes. This limitation becomes clear in Table~\ref{tab:i2v_population_movement}, where the baseline’s population and movement metrics are far worse than the fine-tuned models. 
LoRA improves upon zero-shot but still underperforms the fully fine-tuned model, indicating that updating all parameters more comprehensively aligns the model with domain-specific temporal dynamics. Overall, full fine-tuning yields the most realistic sequences, supporting stable cell proliferation, division, and migration behaviors beyond the initial conditioning frame.

\begin{table}[h]
    \centering
    \caption{Morphological Wasserstein distances for I2V generated sequences}
    \vspace{1em}
    \label{tab:i2v_morphology}
    \begin{tabular}{lcccc}
        \toprule
        \textbf{Model} & \textbf{Area} & \textbf{Eccentricity} & \textbf{Solidity} & \textbf{Perimeter} \\
        \midrule
        Zero-Shot CogVideoX & 76.07 & 0.066 & 0.0098 & 6.36 \\
        LoRA (Rank 256) & 91.36 & 0.034 & 0.0029 & 13.55 \\
        Full Fine-Tune & \textbf{42.92} & \textbf{0.026} & \textbf{0.0026} & \textbf{4.71} \\
        \bottomrule
    \end{tabular}
\end{table}

\begin{table}[ht!]
    \centering
    \caption{Population and movement Wasserstein distances for I2V generated sequences.}
    \vspace{1em}
    \label{tab:i2v_population_movement}
    \begin{tabular}{lcccc}
        \toprule
        \textbf{Model} & \textbf{Final Count} & \textbf{Growth Ratio} & \textbf{Division Events} & \textbf{Net Displacement} \\
        \midrule
        CogVideoX & 178.18 & 2.18 & 36.82 & 35.27 \\
        LoRA Fine-Tune & \textbf{82.09} & 2.79 & 17.26 & 4.55 \\
        Full Fine-Tune & 85.23 & \textbf{0.76} & \textbf{8.21} & \textbf{1.03} \\
        \bottomrule
    \end{tabular}
    \label{tab:i2v_population_movement}
\end{table}

\begin{figure}[ht!]
\centering
\begin{tabular}{c@{\hspace{3mm}}c@{\hspace{2mm}}c@{\hspace{2mm}}c@{\hspace{2mm}}c}
 & Frame 0 & Frame 20 & Frame 60 & Frame 80 \\[2mm]
\raisebox{4ex}{\textbf{CogVideoX}} &
\includegraphics[width=0.20\textwidth]{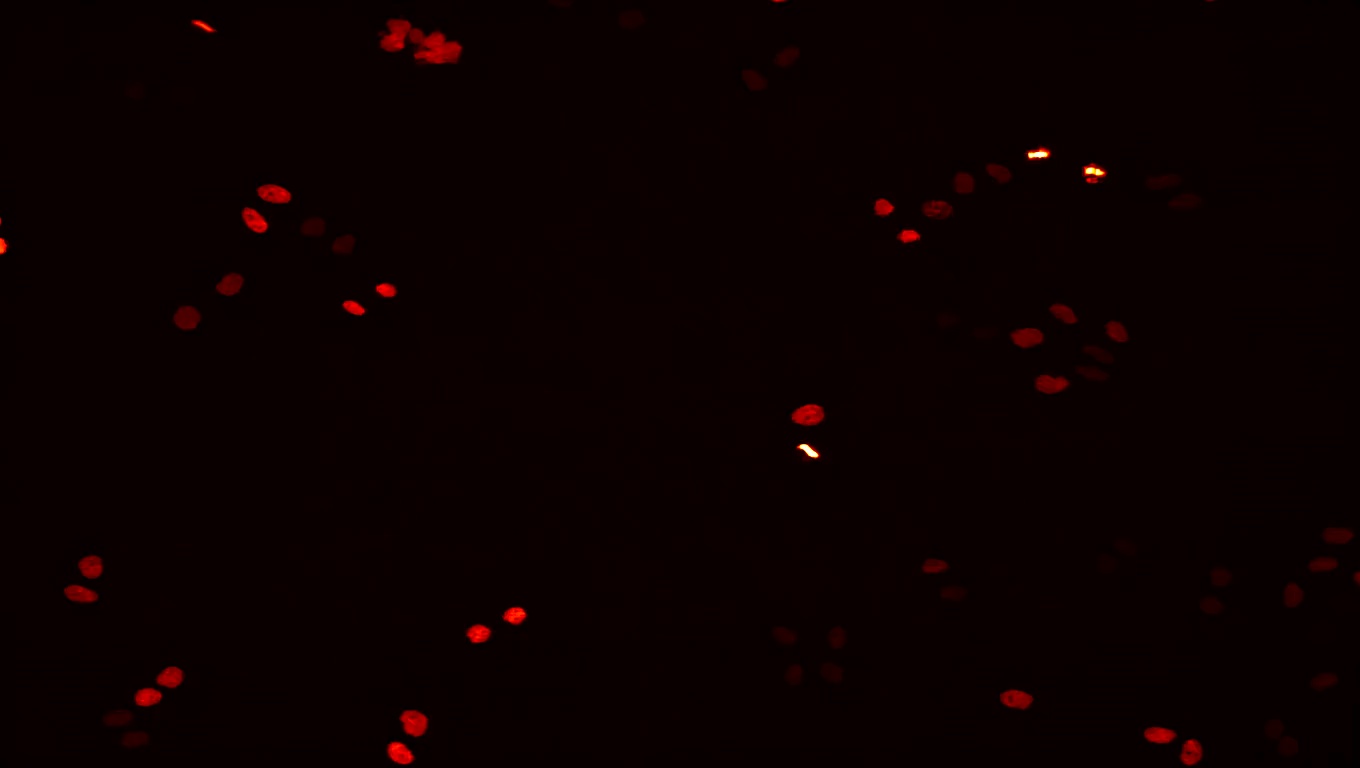} &
\includegraphics[width=0.20\textwidth]{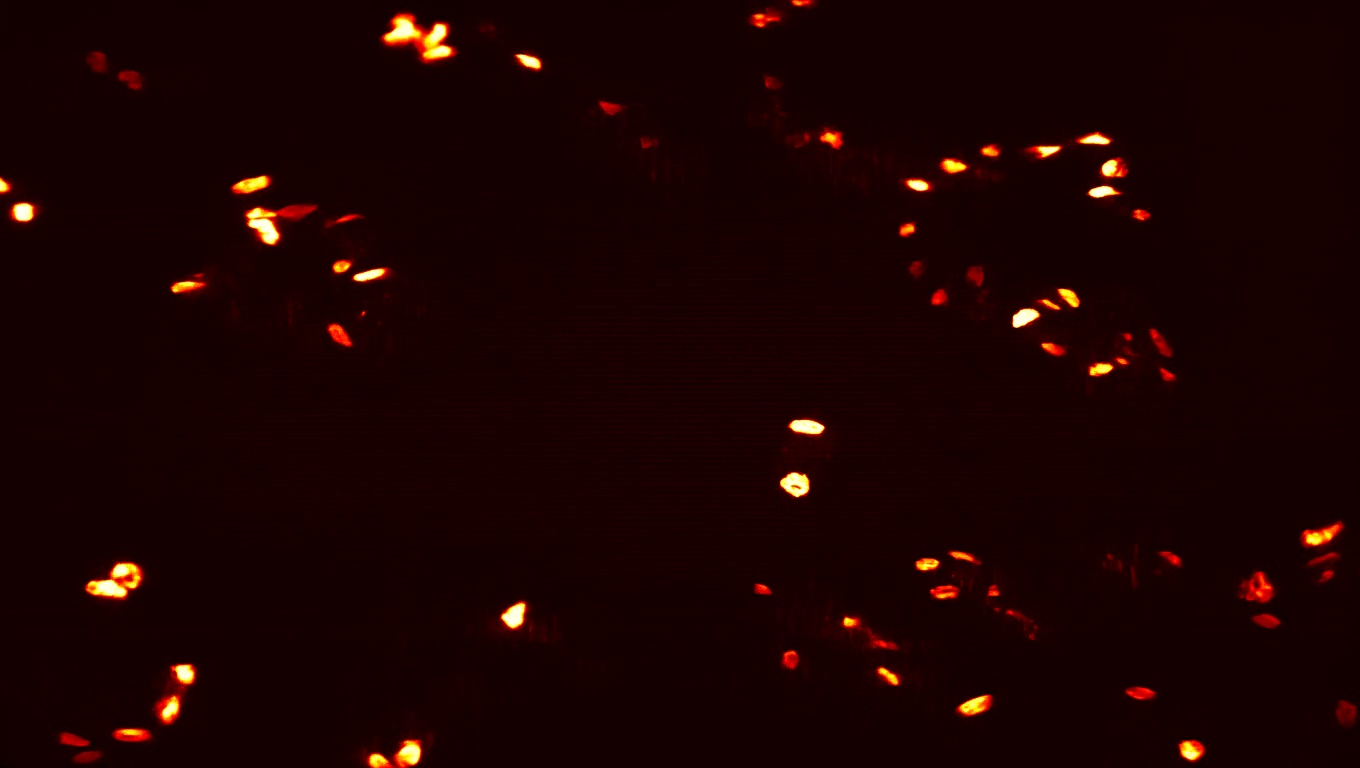} &
\includegraphics[width=0.20\textwidth]{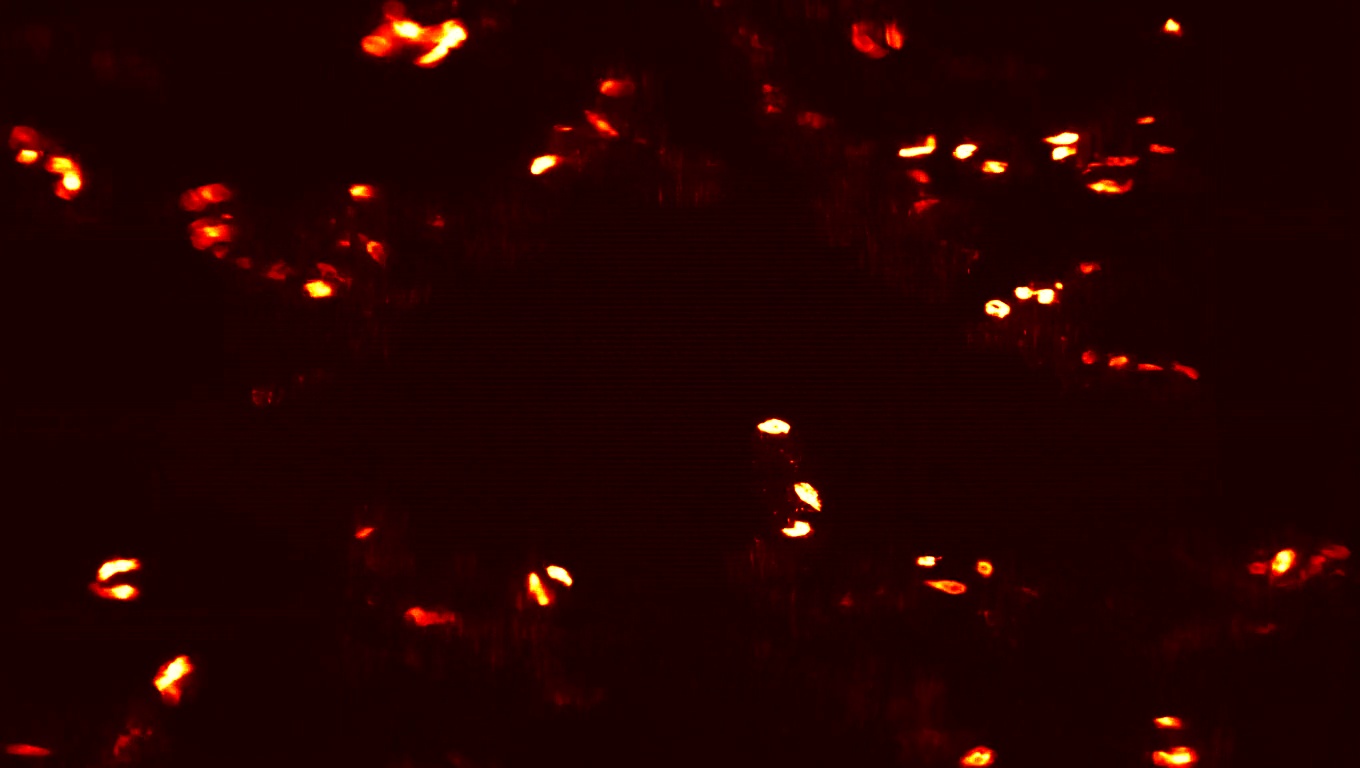} &
\includegraphics[width=0.20\textwidth]{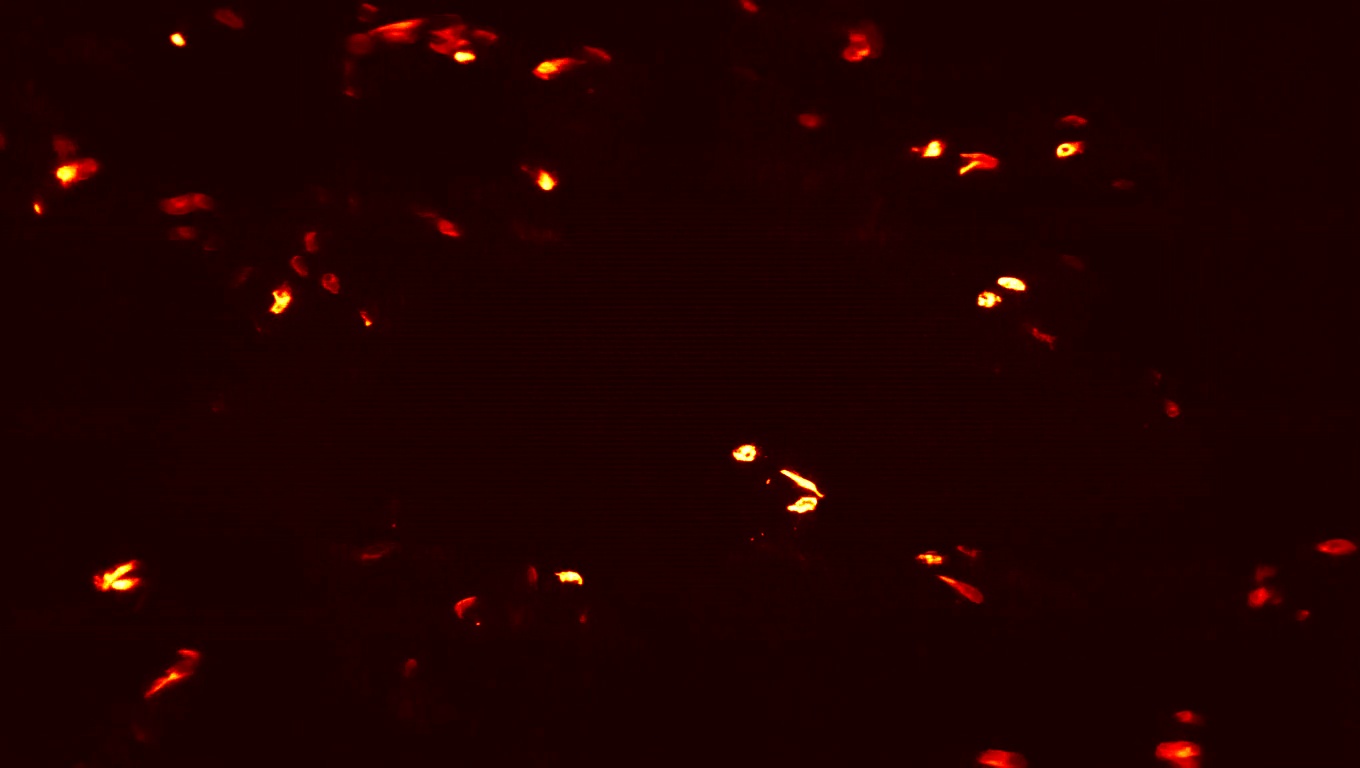} \\[1mm]
\raisebox{4ex}{\textbf{LoRA FT}} &
\includegraphics[width=0.20\textwidth]{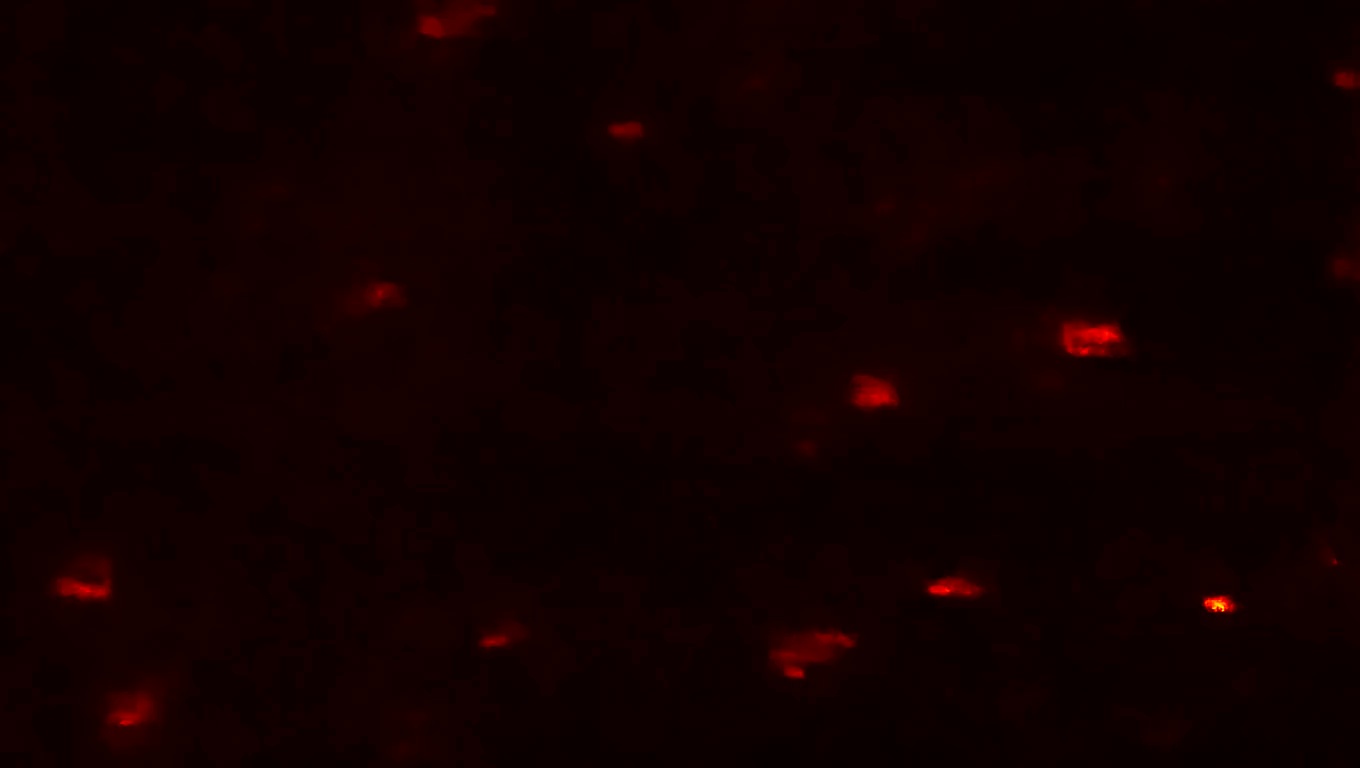} &
\includegraphics[width=0.20\textwidth]{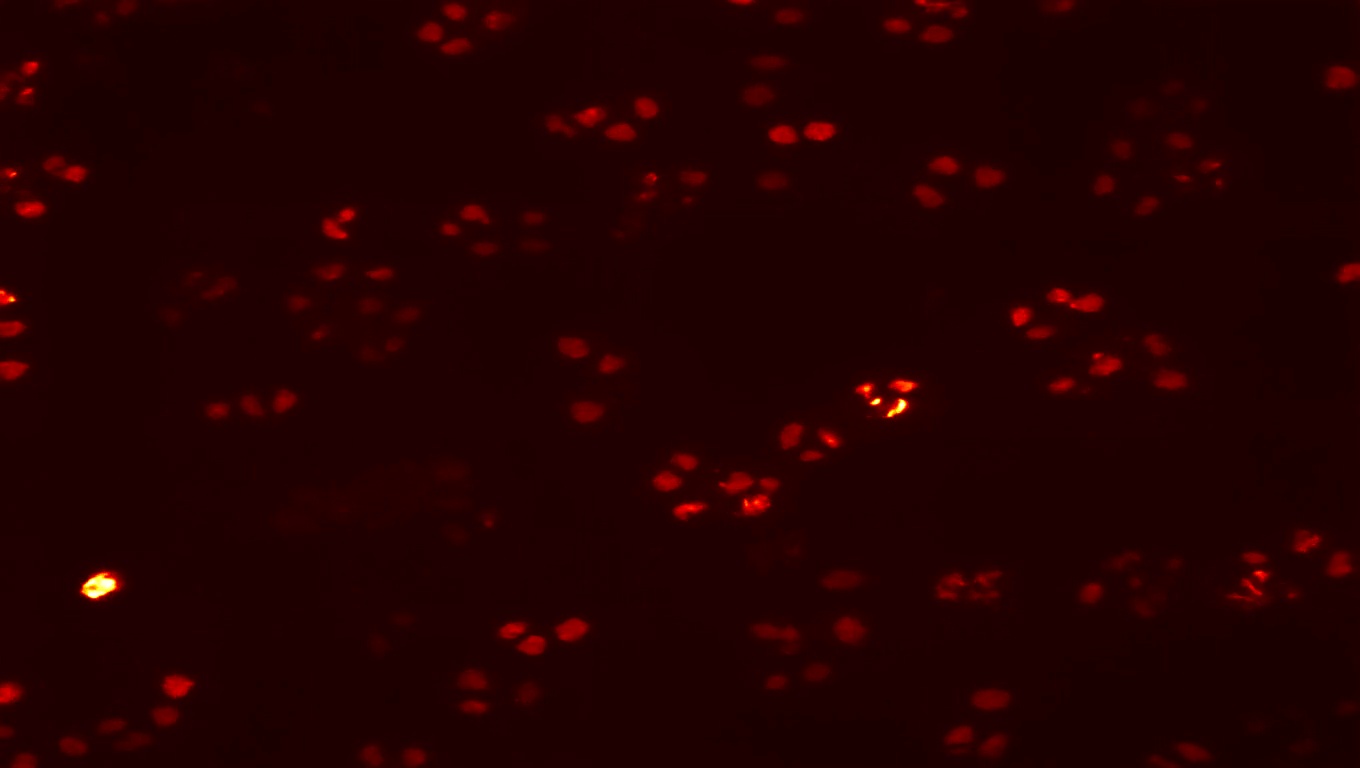} &
\includegraphics[width=0.20\textwidth]{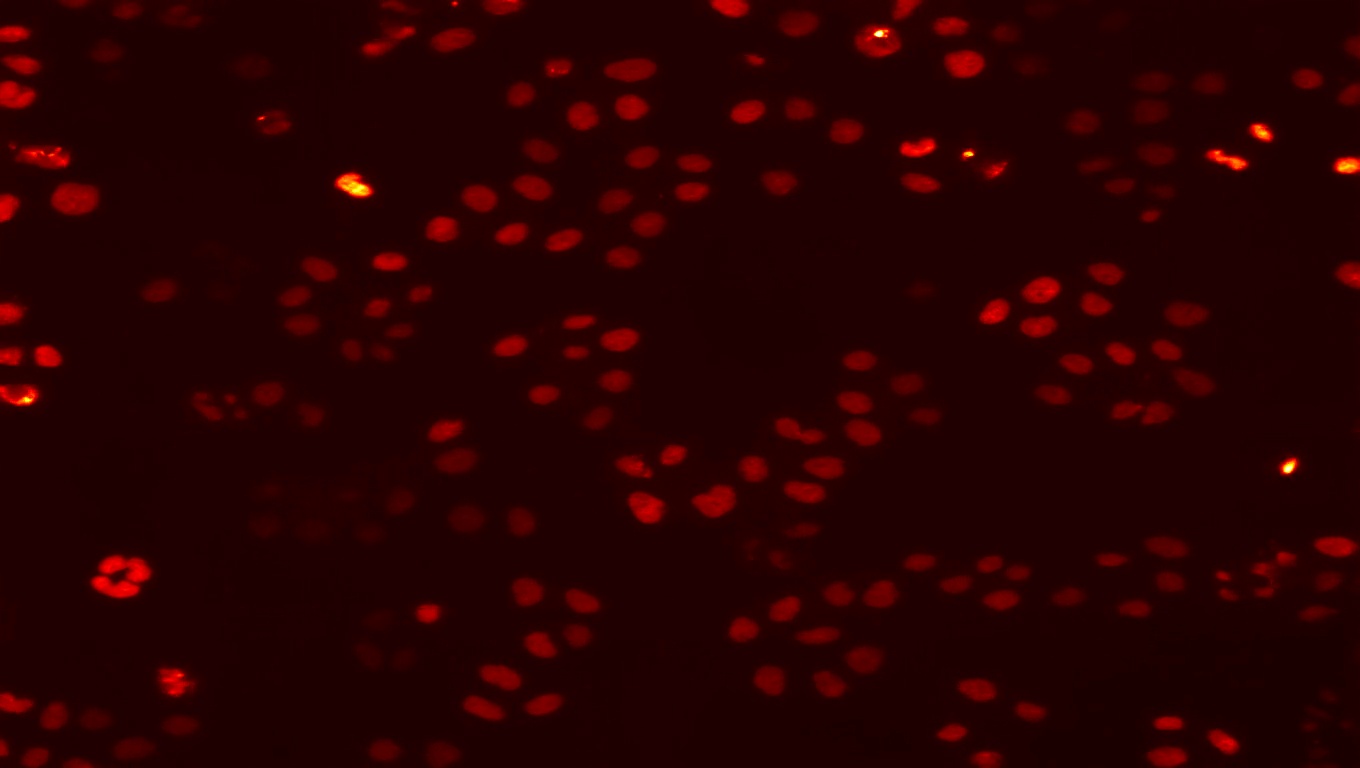} &
\includegraphics[width=0.20\textwidth]{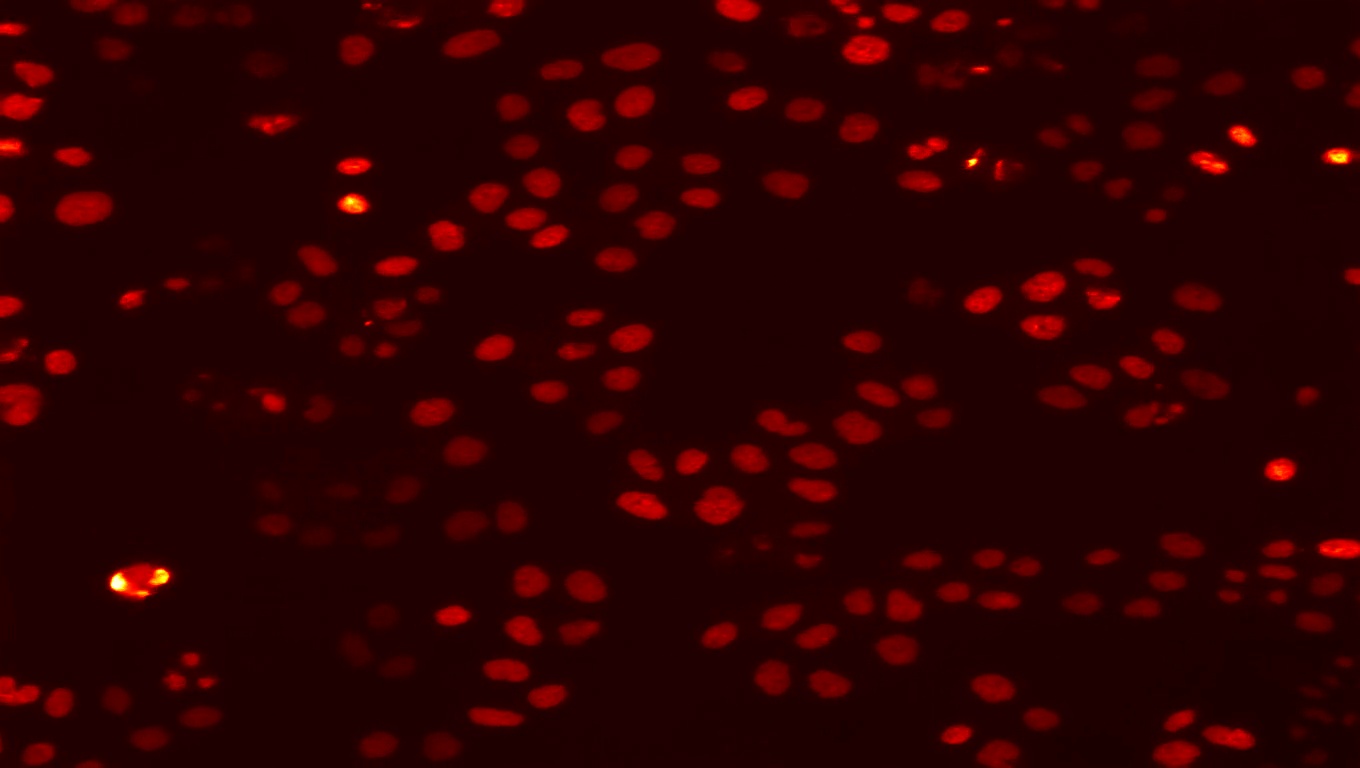} \\[1mm]
\raisebox{4ex}{\textbf{Full FT}} &
\includegraphics[width=0.20\textwidth]{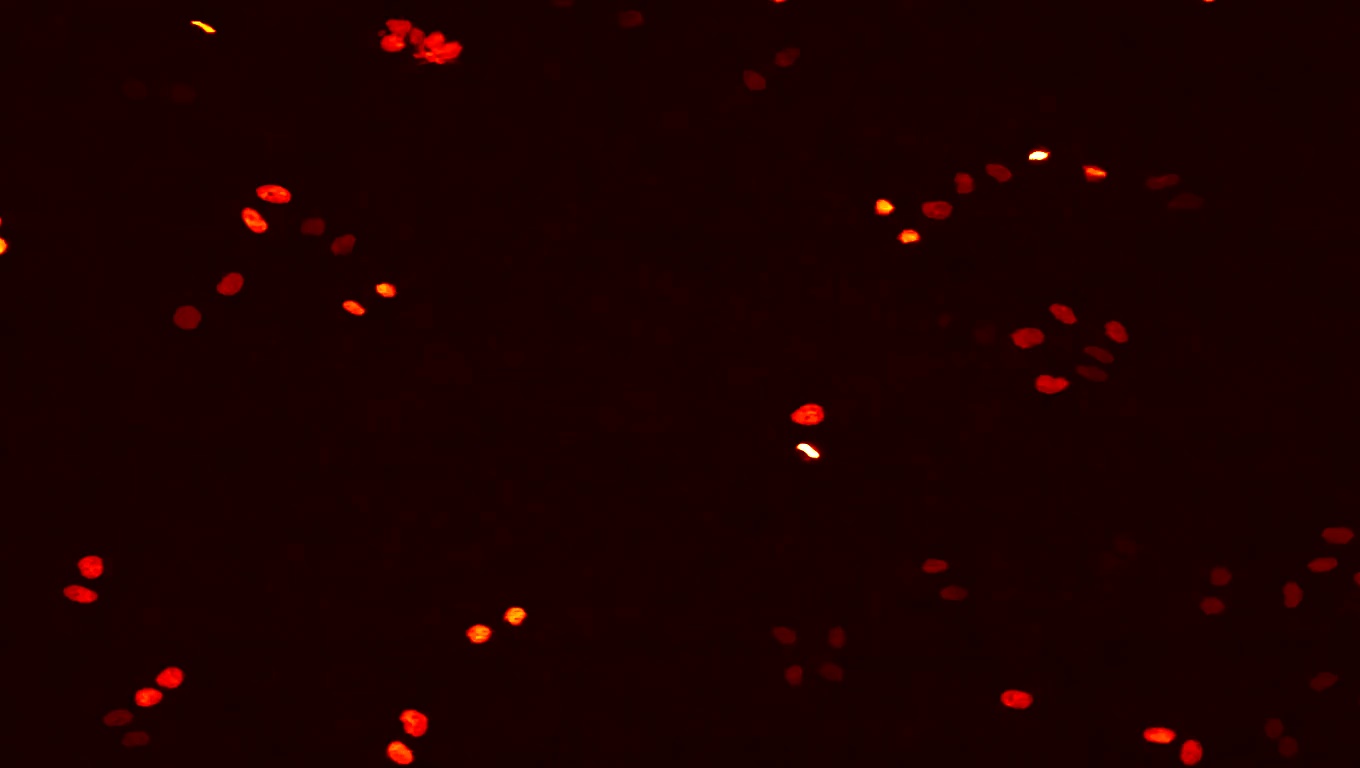} &
\includegraphics[width=0.20\textwidth]{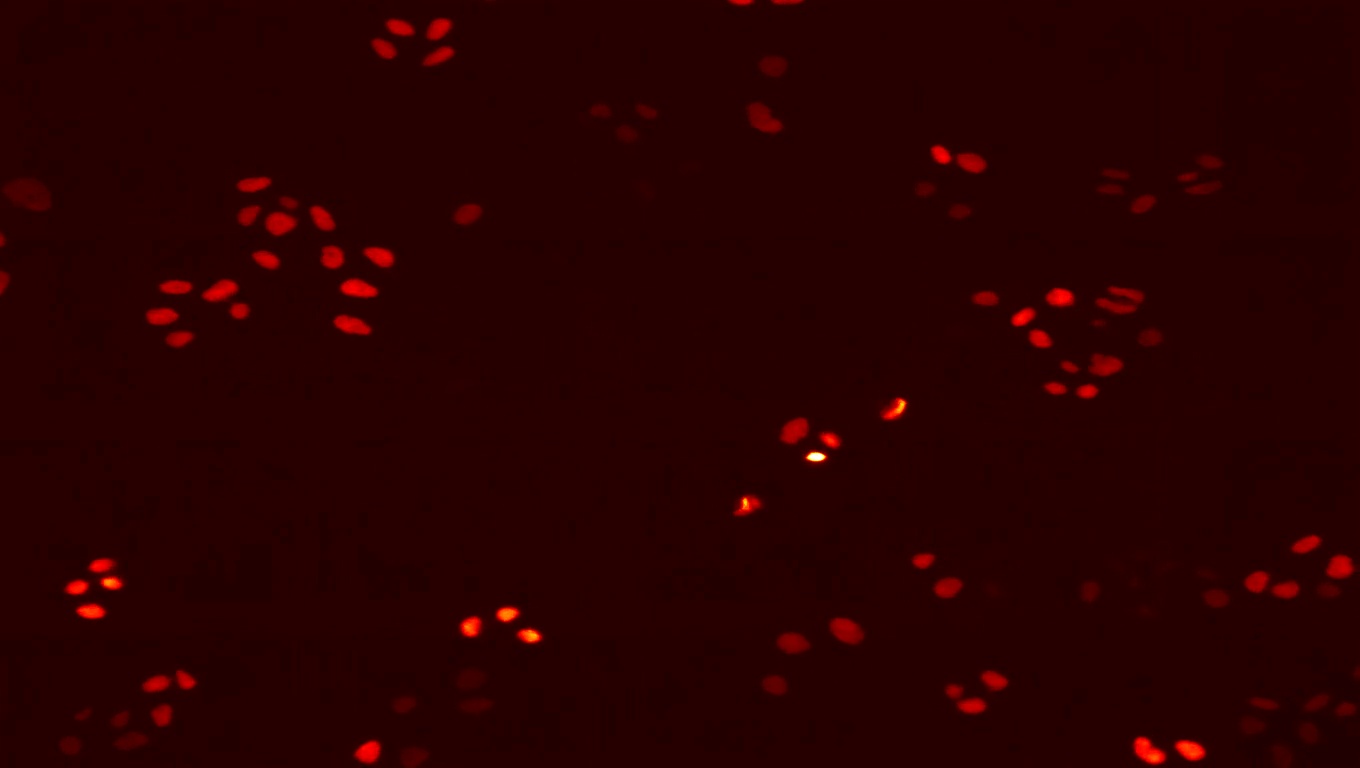} &
\includegraphics[width=0.20\textwidth]{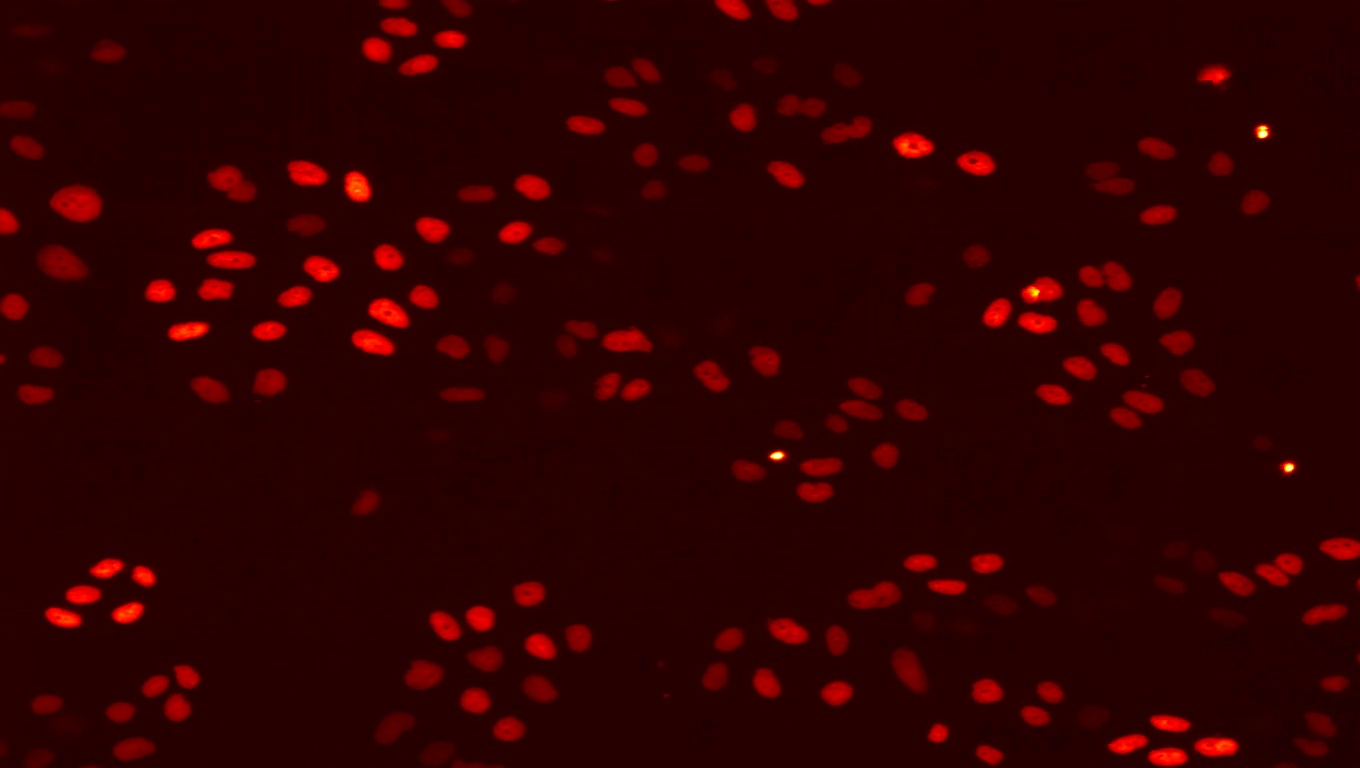} &
\includegraphics[width=0.20\textwidth]{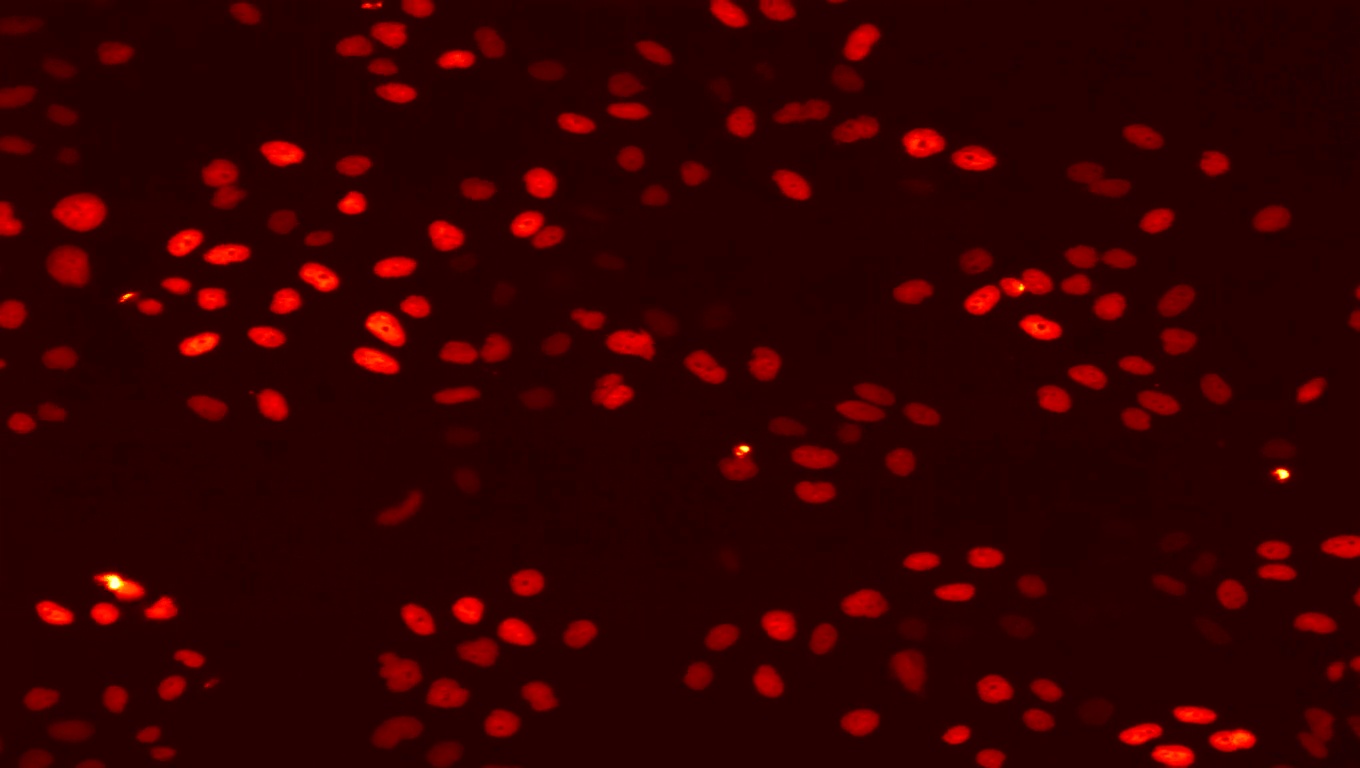} \\[1mm]
\raisebox{4ex}{\textbf{Real}} &
\includegraphics[width=0.20\textwidth]{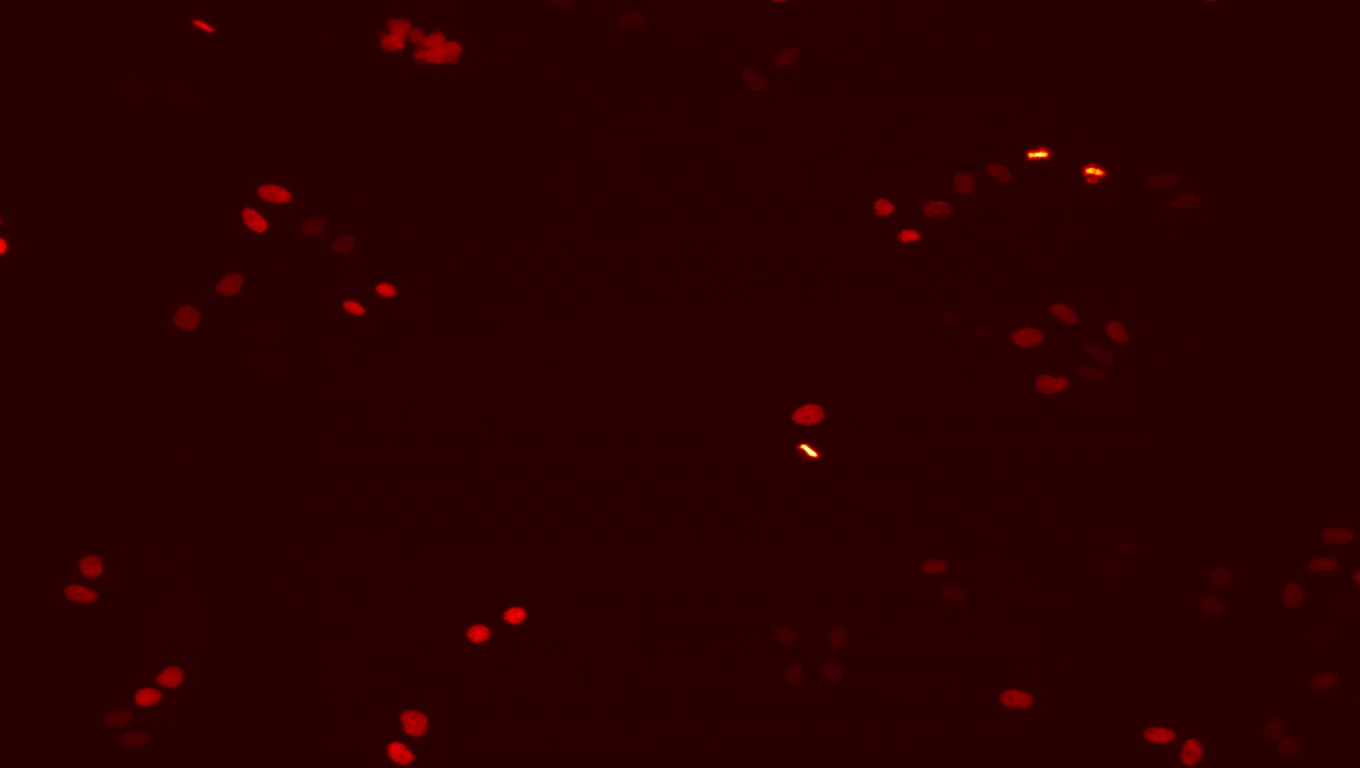} &
\includegraphics[width=0.20\textwidth]{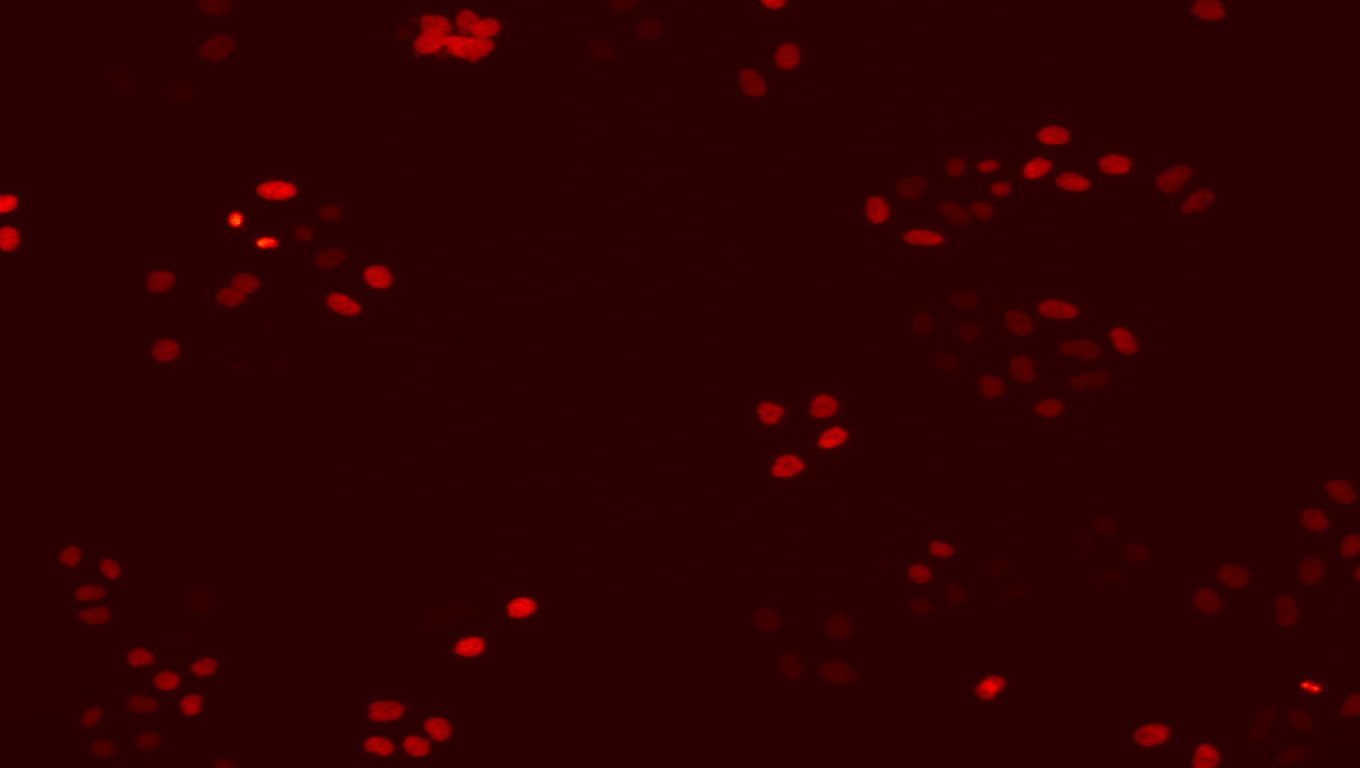} &
\includegraphics[width=0.20\textwidth]{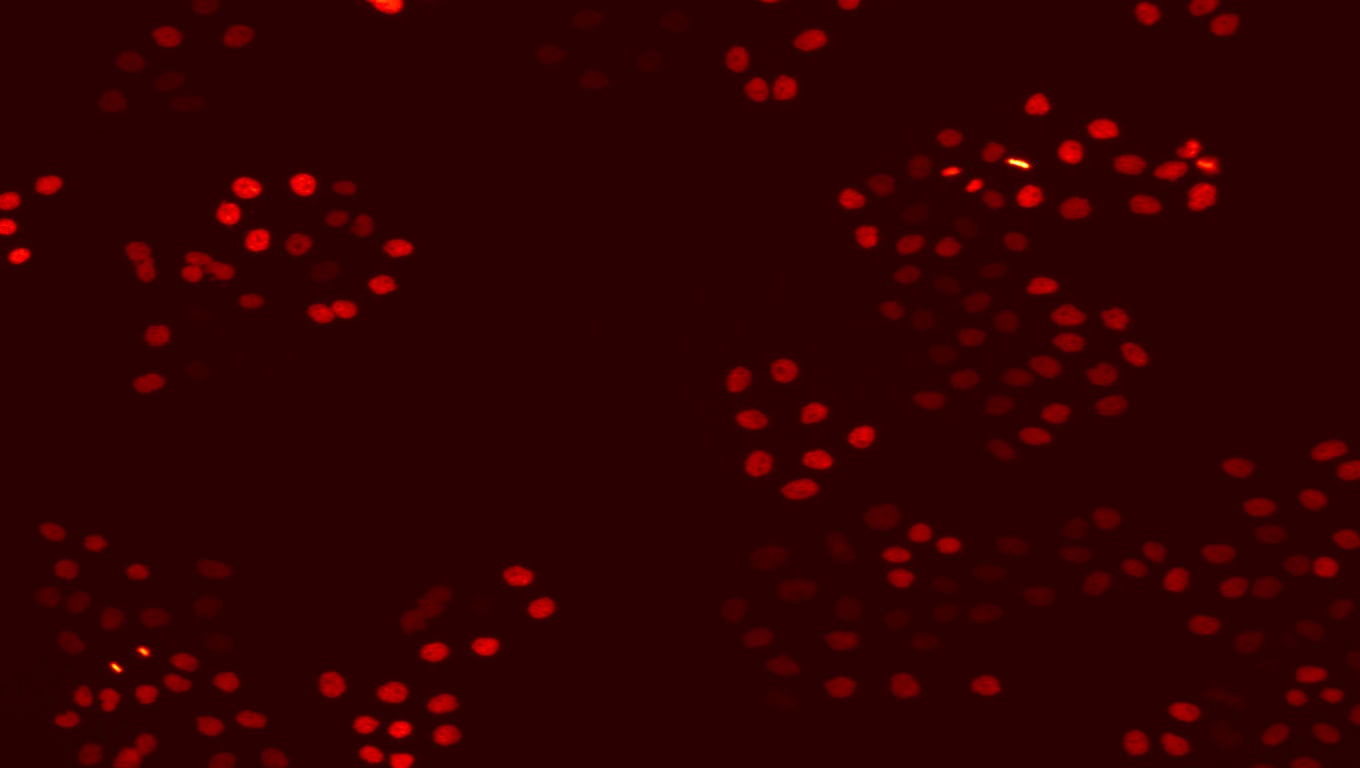} &
\includegraphics[width=0.20\textwidth]{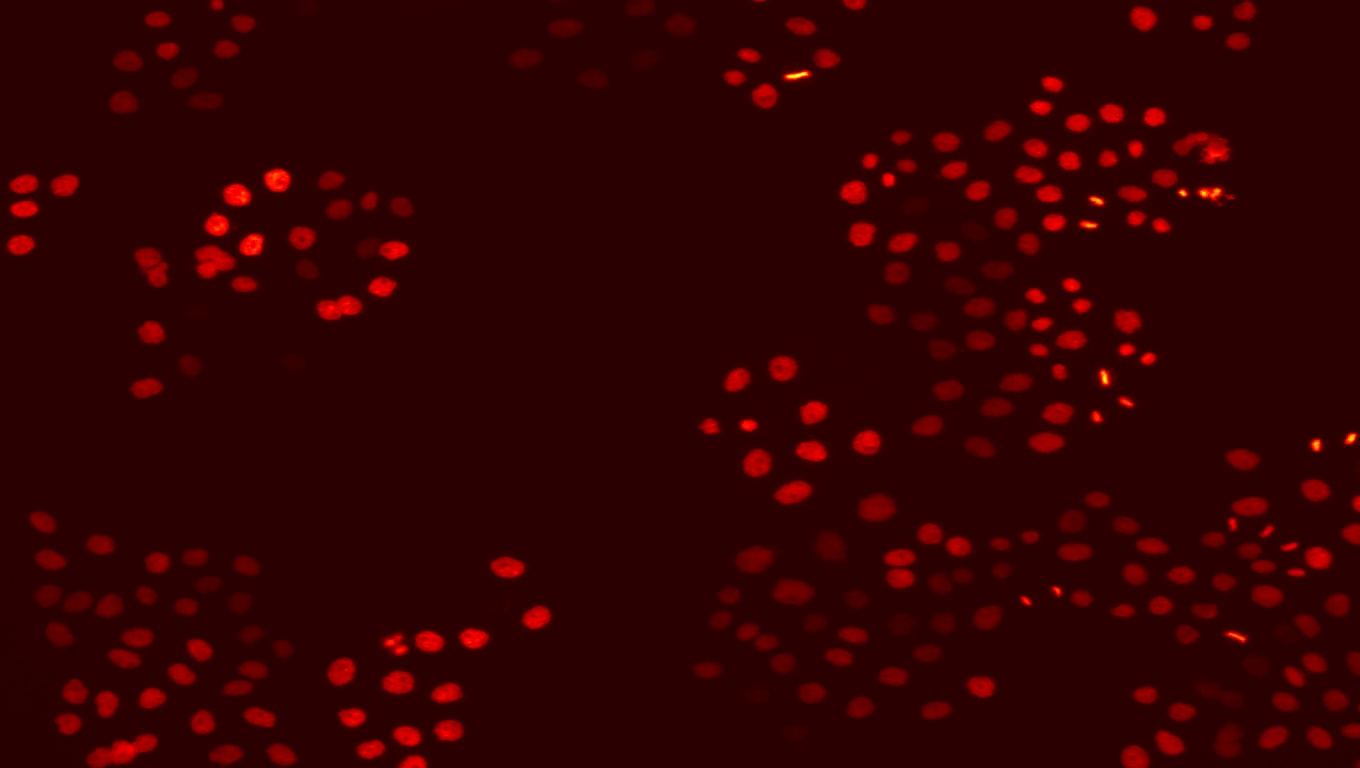} \\
\end{tabular}
\caption{
Visual comparison of zero-shot baseline vs. LoRA fine-tuned vs. full fine-tuned vs. real microscopy sequences. The fine-tuned models generates biologically plausible cell divisions similar to real data, while the zero-shot baseline clearly doesn't.
}
\label{fig:vis_samples3}
\end{figure}

\section{Discussion}
We have shown that fine-tuning large video diffusion models on time-lapse microscopy data yields substantial improvements in visual fidelity and temporal coherence compared to  zero-shot baselines. Our results underscore the unique challenges posed by live-cell imaging: unlike static microscopy datasets, each video must capture 
morphological changes and population-level events over extended periods. 
Although our fine-tuned approach reproduces many of these dynamics 
(even beyond its 81-frame training horizon), conditioning on explicit phenotypes 
(e.g., \emph{HIGH} proliferation or cell death) did not produce sharply differentiated behaviors. This outcome suggests that subtle cell behaviors may require more powerful conditioning mechanisms or larger labeled datasets 
to learn phenotype-specific dynamics. Another constraint stems from the limited availability of extensive, high-quality \emph{live-cell} microscopy datasets for model training and evaluation. While numerous large-scale static microscopy images are publicly available, multi-day time-lapse sequences remain hard to find. Increasing the diversity of time-lapse collections, covering 
a range of genetic and chemical perturbations, would likely improve the model’s ability to capture and discriminate a broader spectrum of cellular states.

\section{Future Work}

To further enhance the realism, specificity, and practical utility of generated time-lapse microscopy videos, several promising directions emerge naturally from our work. First, extending our conditioning approach beyond phenotype scores toward genotype-level information or specific chemical perturbations represents an exciting next step. By directly embedding gene knockdowns or chemical treatment annotations into the model, it might be possible to achieve finer control and enable the model to learn explicit mappings from genetic or chemical perturbations to cellular behaviors.

Another critical direction involves increasing the scale and diversity of available training datasets. While our current experiments rely on a carefully curated subset of microscopy sequences, expanding to larger and more diverse datasets—particularly those spanning various genetic perturbations, different cell lines, and chemical treatments—could enhance the robustness and flexibility of the model. This diversity may also strengthen the model’s ability to generalize to novel scenarios beyond the training distribution. Lastly, refining our conditioning strategies remains an important avenue for future research. Investigating alternative embedding methods, exploring multi-modal fusion techniques that combine text, numeric, and visual conditioning, or introducing hierarchical conditioning mechanisms might enhance the interpretability and controllability of the generated outputs. These methodological refinements hold potential to overcome the current limitations in precisely aligning generated phenotypes with the intended conditioning signals.

A possible application of our approach is in in-silico hypothesis testing, where researchers can systematically generate and analyze cell behaviors (e.g., under different perturbations or phenotypes) without the logistical and financial overhead of repeated lab experiments. By simulating various conditions, like high proliferation or frequent cell death, our time-lapse diffusion models could help biologists explore how different knockdowns or treatments might manifest in practice. Moreover, data augmentation for training downstream tasks (such as nucleus tracking or phenotypic classification) is another avenue where synthetic time-lapse videos can increase diversity in training sets. This is particularly valuable for capturing rare events or underrepresented phenotypes, which might otherwise be too infrequent in real datasets to train robust models.

\section{Data and code availability}

The training dataset is available at \url{https://www.ebi.ac.uk/biostudies/bioimages/studies/S-BIAD865}, and source code is accessible at \url{https://github.com/aicell-lab/cell-video-diffusion}.

\section{Acknowledgments}

We thank Matthew Hartley and Teresa Zulueta Coarasa from the BioImage Archive for recommending the training dataset used in this study. This work is supported by the SciLifeLab \& Wallenberg Data Driven Life Science Program (grant: KAW 2020.0239), the Göran Gustafsson Prize (grant: 2317), and the European Union's Horizon Europe research and innovation program under grant agreement number 101057970 (AI4Life project) awarded to W.O. The computations were enabled by the Berzelius resource provided by the Knut and Alice Wallenberg Foundation at the National Supercomputer Centre in Sweden. We thank Peter Münger from NSC for his support in resource monitoring and allocation on Berzelius. The authors utilized generative AI tools to assist in structuring and drafting portions of this paper.

\bibliographystyle{plain}
\bibliography{main}

\begin{thebibliography}{10}

\bibitem{blattmann2023alignlatentshighresolutionvideo}
Andreas Blattmann, Robin Rombach, Huan Ling, Tim Dockhorn, Seung~Wook Kim, Sanja Fidler, and Karsten Kreis.
\newblock Align your latents: High-resolution video synthesis with latent diffusion models, 2023.

\bibitem{ho2022imagenvideohighdefinition}
Jonathan Ho, William Chan, Chitwan Saharia, Jay Whang, Ruiqi Gao, Alexey Gritsenko, Diederik~P. Kingma, Ben Poole, Mohammad Norouzi, David~J. Fleet, and Tim Salimans.
\newblock Imagen video: High definition video generation with diffusion models, 2022.

\bibitem{ho2022videodiffusionmodels}
Jonathan Ho, Tim Salimans, Alexey Gritsenko, William Chan, Mohammad Norouzi, and David~J. Fleet.
\newblock Video diffusion models, 2022.

\bibitem{hu2021loralowrankadaptationlarge}
Edward~J. Hu, Yelong Shen, Phillip Wallis, Zeyuan Allen-Zhu, Yuanzhi Li, Shean Wang, Lu~Wang, and Weizhu Chen.
\newblock Lora: Low-rank adaptation of large language models, 2021.

\bibitem{kong2025hunyuanvideosystematicframeworklarge}
Weijie Kong, Qi~Tian, Zijian Zhang, Rox Min, Zuozhuo Dai, Jin Zhou, Jiangfeng Xiong, Xin Li, Bo~Wu, Jianwei Zhang, Kathrina Wu, Qin Lin, Junkun Yuan, Yanxin Long, Aladdin Wang, Andong Wang, Changlin Li, Duojun Huang, Fang Yang, Hao Tan, Hongmei Wang, Jacob Song, Jiawang Bai, Jianbing Wu, Jinbao Xue, Joey Wang, Kai Wang, Mengyang Liu, Pengyu Li, Shuai Li, Weiyan Wang, Wenqing Yu, Xinchi Deng, Yang Li, Yi~Chen, Yutao Cui, Yuanbo Peng, Zhentao Yu, Zhiyu He, Zhiyong Xu, Zixiang Zhou, Zunnan Xu, Yangyu Tao, Qinglin Lu, Songtao Liu, Dax Zhou, Hongfa Wang, Yong Yang, Di~Wang, Yuhong Liu, Jie Jiang, and Caesar Zhong.
\newblock Hunyuanvideo: A systematic framework for large video generative models, 2025.

\bibitem{Navidi2024.12.19.629451}
Zeinab Navidi, Jun Ma, Esteban~A. Miglietta, Le~Liu, Anne~E. Carpenter, Beth~A. Cimini, Benjamin Haibe-Kains, and Bo~Wang.
\newblock Morphodiff: Cellular morphology painting with diffusion models.
\newblock {\em bioRxiv}, 2024.

\bibitem{Neumann2010}
Beate Neumann, Thomas Walter, Jean-Karim H{\'e}rich{\'e}, Jutta Bulkescher, Holger Erfle, , et~al.
\newblock Phenotypic profiling of the human genome by time-lapse microscopy reveals cell division genes.
\newblock {\em Nature}, 464(7289):721, 2010.
\newblock hal-01144034.

\bibitem{Stringer2020CellposeAG}
Carsen Stringer, Tim Wang, Michalis Michaelos, and Marius Pachitariu.
\newblock Cellpose: a generalist algorithm for cellular segmentation.
\newblock {\em Nature Methods}, 18:100 -- 106, 2020.

\bibitem{yang2024cogvideoxtexttovideodiffusionmodels}
Zhuoyi Yang, Jiayan Teng, Wendi Zheng, Ming Ding, Shiyu Huang, Jiazheng Xu, Yuanming Yang, Wenyi Hong, Xiaohan Zhang, Guanyu Feng, Da~Yin, Xiaotao Gu, Yuxuan Zhang, Weihan Wang, Yean Cheng, Ting Liu, Bin Xu, Yuxiao Dong, and Jie Tang.
\newblock Cogvideox: Text-to-video diffusion models with an expert transformer, 2024.

\bibitem{yuan2024chronomagicbenchbenchmarkmetamorphicevaluation}
Shenghai Yuan, Jinfa Huang, Yongqi Xu, Yaoyang Liu, Shaofeng Zhang, Yujun Shi, Ruijie Zhu, Xinhua Cheng, Jiebo Luo, and Li~Yuan.
\newblock Chronomagic-bench: A benchmark for metamorphic evaluation of text-to-time-lapse video generation, 2024.

\end{thebibliography}

\clearpage 
\appendix
\section*{Supplementary Material}
\label{sec:supp}
\section*{Appendix A: Training Details}
\vspace{5pt}
\noindent
\textbf{Hardware.}
All experiments were conducted on a single node equipped with four NVIDIA A100 GPUs (80\,GB VRAM each). We employed DeepSpeed ZeRO-2 optimization to effectively handle larger batch sizes while keeping memory usage within practical limits.

\vspace{5pt}
\noindent
\textbf{Hyperparameters.}
Table~\ref{tab:hyperparams} summarizes the key hyperparameters used during training. All models were trained for 1500 steps at a global batch size of 4. Across the entire training run, this corresponds to \textit{6000 samples} processed in total (4 samples per step $\times$ 1500 steps), which is equivalent to approximately 2 epochs.

\begin{table}[h]
\centering
\begin{tabular}{@{}ll@{}}
\toprule
\textbf{Hyperparameter} & \textbf{Value} \\
\midrule
Global batch size       & 4 \\
Learning rate           & $1\times10^{-4}$ \\
Optimizer               & AdamW \\
Training steps          & 1500 \\
LoRA rank (I2V only)    & 256 \\
\bottomrule
\end{tabular}
\caption{Key hyperparameters for our training runs.}
\label{tab:hyperparams}
\end{table}

\vspace{5pt}
\noindent
\textbf{Learning Rate Schedule.}
We employed a warmup phase for the first 100 steps (linearly ramping the learning rate from $0$ to the target value), followed by a \emph{constant} learning rate for the remainder of training.

\vspace{5pt}
\noindent
\textbf{Software.}
PyTorch and Diffusers were used for the training code. All code for training, data preprocessing, and evaluation is publicly available at \url{https://github.com/aicell-lab/cell-video-diffusion}.

\section*{Appendix B: Evaluation Details}

\subsection*{Morphological Metrics}

\begin{algorithm}[H]
\caption{Computing Morphological Descriptors for a Time-Lapse Video}
\label{alg:morphology}
\begin{algorithmic}[1]
\REQUIRE A time-lapse mask array $\mathcal{M} \in \{0,1\}^{T \times H \times W}$, where $T$ is the number of frames.
\ENSURE A set $\mathcal{D}_{\mathrm{morph}}$ of nucleus-level descriptors (e.g.\ area, eccentricity).

\STATE $\mathcal{D}_{\mathrm{morph}} \leftarrow \{\}$ \quad \textit{/* Initialize empty collection */}
\FOR{$t \leftarrow 0$ to $T-1$}
    \STATE $\mathcal{L}_t \leftarrow \text{LabelConnectedComponents}(\mathcal{M}[t])$
    \FOR{each connected component $c$ in $\mathcal{L}_t$}
        \STATE \textit{area} $\leftarrow \text{Size}(c)$ 
        \STATE \textit{eccentricity} $\leftarrow \text{Eccentricity}(c)$ 
        \STATE \textit{solidity} $\leftarrow \text{Solidity}(c)$ 
        \STATE \textit{perimeter} $\leftarrow \text{Perimeter}(c)$ 
        \STATE $\mathcal{D}_{\mathrm{morph}}.\text{add}(\{\textit{area},\textit{eccentricity},\textit{solidity},\textit{perimeter}\})$
    \ENDFOR
\ENDFOR
\RETURN $\mathcal{D}_{\mathrm{morph}}$
\end{algorithmic}
\end{algorithm}

\subsection*{Movement Metrics}

\begin{algorithm}[H]
\caption{Computing Movement Metrics for a Time-Lapse Video}
\label{alg:movement}
\begin{algorithmic}[1]
\REQUIRE A time-lapse mask array $\mathcal{M} \in \{0,1\}^{T \times H \times W}$, where $T$ is the number of frames.
\ENSURE A set of per-nucleus movement metrics (e.g., speed, total distance, displacement).

\STATE \textbf{Initialize} an empty set of \textit{tracks}, each track capturing the positions of a single nucleus over time.

\vspace{0.25em}
\STATE \textbf{Step 1: Centroid Extraction}
\FOR{$t \leftarrow 0$ to $T-1$}
    \STATE Identify connected components in $\mathcal{M}[t]$ (one per nucleus).
    \STATE Extract the centroid $(y,x)$ of each nucleus. 
    \STATE Assign each nucleus in frame $t$ to either an \emph{existing} track or create a \emph{new} track, typically by a nearest-centroid matching to the previous frame's nuclei.
\ENDFOR

\vspace{0.25em}
\STATE \textbf{Step 2: Compute Frame-to-Frame Movement}
\FOR{each track $k$ in \textit{tracks}}
    \IF{\emph{track} $k$ has $n_k$ frames ($n_k < 2$)}
        \STATE Skip (no movement data).
    \ELSE
        \STATE Let $(y_1,x_1), \dots, (y_{n_k}, x_{n_k})$ be the centroids in time order.
        \STATE Compute \textit{displacements}:
        \[
          d_i = \sqrt{(y_{i+1}-y_i)^2 + (x_{i+1}-x_i)^2}\quad \text{for}\ i=1,\dots,n_k{-}1.
        \]
        \STATE Sum these to get \textit{total distance} $D = \sum d_i$.
        \STATE \textit{Net displacement} $D_{\mathrm{net}} = \| (y_{n_k},x_{n_k}) - (y_1,x_1) \|_2$.
        \STATE \textit{Average speed} $v_{\mathrm{avg}} = D / (n_k - 1)$.
        \STATE \textit{Directness ratio} $= D_{\mathrm{net}} / D$.
        \STATE Store these metrics in the \emph{track record}.
    \ENDIF
\ENDFOR

\vspace{0.25em}
\STATE \textbf{Return} an aggregate of per-track movement metrics (e.g., average speed, total distance).
\end{algorithmic}
\end{algorithm}

\subsection*{Population Metrics}

\begin{algorithm}[H]
\caption{Computing Population-Level Statistics for a Time-Lapse Video}
\label{alg:population}
\begin{algorithmic}[1]
\REQUIRE A time-lapse mask array $\mathcal{M} \in \{0,1\}^{T \times H \times W}$, where $T$ is the number of frames.
\ENSURE A set of population metrics (e.g., cell counts, growth ratio, division events).

\STATE \textbf{Initialize:} $\mathit{counts} \gets \varnothing$ \quad /* store nuclei counts per frame */
\FOR{$t \leftarrow 0$ to $T-1$}
    \STATE $\mathit{counts}[t] \leftarrow \text{NumConnectedComponents}(\mathcal{M}[t])$ 
          /* number of distinct nuclei */
\ENDFOR
\STATE $\mathit{initialCount} \leftarrow \mathit{counts}[0]$
\STATE $\mathit{finalCount} \leftarrow \mathit{counts}[T-1]$
\STATE $\mathit{growthRatio} \leftarrow \frac{\mathit{finalCount}}{\max(\mathit{initialCount},1)}$
\STATE $\mathit{growthAbsolute} \leftarrow \mathit{finalCount} - \mathit{initialCount}$

\vspace{1ex}
\STATE /* \textit{Division detection via derivative thresholding (optional):} */
\STATE $\mathit{derivative}[t] \gets \mathit{counts}[t{+}1] - \mathit{counts}[t] \;\;\;\; \forall\, t \in [0\ldots T{-}2]$
\STATE $\mathit{divisionFrames} \gets \{\,t \mid \mathit{derivative}[t] > \tau\}$ 
      /* e.g. $\tau = 0.5$ for a “significant” cell-count jump */
\STATE $\mathit{divisionCount} \gets \bigl|\mathit{divisionFrames}\bigr|$
\IF{$\mathit{divisionCount} \ge 2$}
    \STATE $\mathit{avgDivisionInterval} \gets \text{mean}\bigl(\Delta(\mathit{divisionFrames})\bigr)$
\ELSE
    \STATE $\mathit{avgDivisionInterval} \gets \text{NaN}$
\ENDIF

\RETURN $\{\,
  \mathit{initialCount},\,
  \mathit{finalCount},\,
  \mathit{growthRatio},\,
  \mathit{growthAbsolute},\,
  \mathit{divisionCount},\,
  \mathit{avgDivisionInterval}
\}$
\end{algorithmic}
\end{algorithm}

\noindent
These procedures yields a distribution of movement metrics over all tracked nuclei. To compare real vs.\ generated videos, we form two distributions of movement metrics and measure their discrepancy via the Wasserstein distance $W_{1}$: 
\[
W_1(F, G)
\,=\,
\int_0^1
\Bigl|F^{-1}(u) \;-\; G^{-1}(u)\Bigr|
\,du,
\]
where $F$ and $G$ are the respective cumulative distribution functions of the chosen descriptor. 
A lower $W_{1}$ indicates a closer match between real and synthetic distributions for that morphology metric. 
We repeat this process frame-by-frame or globally (across all frames) depending on the level of analysis desired.

\end{document}


\section*{Appendix A: Training Details}
\vspace{5pt}
\noindent
\textbf{Hardware.}
All experiments were conducted on a single node equipped with four NVIDIA A100 GPUs (80\,GB VRAM each). We employed DeepSpeed ZeRO-2 optimization to effectively handle larger batch sizes while keeping memory usage within practical limits.

\vspace{5pt}
\noindent
\textbf{Hyperparameters.}
Table~\ref{tab:hyperparams} summarizes the key hyperparameters used during training. All models were trained for 1500 steps at a global batch size of 4. Across the entire training run, this corresponds to \textit{6000 samples} processed in total (4 samples per step $\times$ 1500 steps), which is equivalent to approximately 2 epochs.

\begin{table}[h]
\centering
\begin{tabular}{@{}ll@{}}
\toprule
\textbf{Hyperparameter} & \textbf{Value} \\
\midrule
Global batch size       & 4 \\
Learning rate           & $1\times10^{-4}$ \\
Optimizer               & AdamW \\
Training steps          & 1500 \\
LoRA rank (I2V only)    & 256 \\
\bottomrule
\end{tabular}
\caption{Key hyperparameters for our training runs.}
\label{tab:hyperparams}
\end{table}

\vspace{5pt}
\noindent
\textbf{Learning Rate Schedule.}
We employed a warmup phase for the first 100 steps (linearly ramping the learning rate from $0$ to the target value), followed by a \emph{constant} learning rate for the remainder of training.

\vspace{5pt}
\noindent
\textbf{Software.}
PyTorch and Diffusers were used for the training code. The codebase will be open source soon.

\section*{Appendix B: Supplementary Videos}

In this section, we provide various supplementary videos to illustrate real microscopy data (with different phenotypic conditions) and our fine-tuned diffusion model’s outputs across both text-to-video (T2V) and image-to-video (I2V) settings. Each category lists the relevant clips, which reviewers may view alongside the main paper’s descriptions of experimental conditions.

\vspace{8pt}
\noindent
\textbf{1. Real Videos.}
\begin{itemize}
    \item \texttt{real\_high\_cell\_count.mp4}
    \item \texttt{real\_high\_cell\_death.mp4}
    \item \texttt{real\_high\_proliferation.mp4}
    \item \texttt{real\_low\_cell\_count.mp4}
    \item \texttt{real\_low\_cell\_death.mp4}
    \item \texttt{real\_low\_proliferation.mp4}
    \item \texttt{real\_video0.mp4}, \texttt{real\_video1.mp4}, \texttt{real\_video2.mp4}
\end{itemize}
These clips illustrate ground-truth microscopy time-lapses under various phenotypic conditions (\emph{e.g.}, higher vs.\ lower rates of cell death or proliferation). They serve as the reference standard against which we compare our generated sequences. Cells dying almost looks like flickering artifacts (as can be seen in \texttt{real\_high\_cell\_death.mp4}), so when observing flickering in generated videos, that is not an error with the model, it is the visualization of cell death! 

\vspace{8pt}
\noindent
\textbf{2. Text-to-Video (T2V) Outputs.}
\begin{itemize}
    \item \texttt{text2video\_baseline.mp4}: T2V generation results from the raw CogVideoX 1.5 model.
    \item \texttt{text2video\_generated0.mp4} 
    \item \texttt{text2video\_generated1.mp4} 
    \item \texttt{text2video\_generated2.mp4}
\end{itemize}

\vspace{8pt}
\noindent
\textbf{3. Image-to-Video (I2V) Outputs.}
\begin{itemize}
    \item \texttt{image2video\_baseline.mp4}: I2V generation results from the raw CogVideoX 1.5 model.
    \item \texttt{image2video\_generated0.mp4}
    \item \texttt{image2video\_generated1.mp4}
    \item \texttt{image2video\_generated2.mp4}
\end{itemize}

\vspace{8pt}
\noindent
\textbf{4. Extended Sequence Generation.}
\begin{itemize}
    \item \texttt{generated\_129frames0.mp4}
    \item \texttt{generated\_129frames1.mp4}
    \item \texttt{generated\_129frames2.mp4}
\end{itemize}
These examples demonstrates that our fine-tuned model can coherently generate cell dynamics over longer horizons (129 frames) despite being trained primarily on 81-frame sequences.